\documentclass{article}

 \PassOptionsToPackage{numbers, compress}{natbib}
\usepackage[final]{nips_2018}

\usepackage[utf8]{inputenc} % allow utf-8 input
\usepackage[T1]{fontenc}    % use 8-bit T1 fonts
\usepackage{hyperref}       % hyperlinks
\usepackage{url}            % simple URL typesetting
\usepackage{booktabs}       % professional-quality tables
\usepackage{amsfonts}       % blackboard math symbols
\usepackage{nicefrac}       % compact symbols for 1/2, etc.
\usepackage{microtype}      % microtypography

\usepackage{enumitem}
\usepackage{bm}
\usepackage{chngpage}
\usepackage{times}
\usepackage{amsmath}
\usepackage{amssymb}
\usepackage{wrapfig}
\usepackage{color, colortbl}
\usepackage[english]{babel}
\usepackage{mathrsfs}
\usepackage{tikz}
\usetikzlibrary{bayesnet}
\usepackage{pgfplots}
\usepackage{cleveref}
\usepackage{subcaption}
\usepackage{algorithm}
\usepackage{algorithmic}
\usepackage{multirow}
\usepackage{makecell}
\usepackage{adjustbox}
\usepackage{verbatimbox}
\newcommand\Includegraphics[2][]{\addvbuffer[2pt 0pt]{\includegraphics[#1]{#2}}}

\newcommand{\bo}{\mathbf{o}}
\newcommand{\bx}{\mathbf{x}}

\newcommand{\bz}{\mathbf{z}}
\newcommand{\bbm}{\mathbf{m}}

\newcommand{\bt}{\mathbf{t}}

\newcommand{\bL}{\mathbf{L}}

\newcommand{\bS}{\mathbf{S}}
\newcommand{\bO}{\mathbf{O}}
\newcommand{\bX}{\mathbf{X}}
\newcommand{\bZ}{\mathbf{Z}}
\newcommand{\balpha}{{\bm{\alpha}}}
\newcommand{\bbeta}{{\bm{\beta}}}
\newcommand{\bnu}{{\bm{\nu}}}
\newcommand{\bmeta}{{\bm{\eta}}}
\newcommand{\beps}{\bm{\epsilon}}
\newcommand{\bpsi}{{\bm{\psi}}}
\newcommand{\bphi}{{\bm{\phi}}}
\newcommand{\bpi}{{\bm{\pi}}}
\newcommand{\bmu}{{\bm{\mu}}}
\newcommand{\bsigma}{{\bm{\sigma}}}

\newcommand{\bxi}{\bm{\xi}}
\newcommand{\btheta}{{\bm{\theta}}}
\newcommand{\bTheta}{{\bm{\Theta}}}
\newcommand{\bSigma}{{\bm{\Sigma}}}
\newcommand{\bGamma}{{\bm{\Gamma}}}
\newcommand{\bgamma}{{\bm{\gamma}}}

\DeclareMathOperator{\KL}{KL}
\newcommand{\opt}[1]{{#1}^{*} \!}

\title{Semi-crowdsourced Clustering with\\Deep Generative Models}

\author{
    Yucen Luo,~ Tian Tian,~ Jiaxin Shi,~ Jun Zhu\thanks{corresponding author},~ Bo Zhang\\
    Dept. of Comp. Sci. \& Tech., Institute for AI, THBI Lab, BNRist Center,\\
    State Key Lab for Intell. Tech. \& Sys., Tsinghua University, Beijing, China\\	
    \texttt{\{luoyc15,shijx15\}@mails.tsinghua.edu.cn, rossowhite@163.com}\\
    \texttt{\{dcszj,dcszb\}@mail.tsinghua.edu.cn}
}

\begin{document}
    % \nipsfinalcopy is no longer used
    
    \maketitle
    
    \vspace{-0.5cm}
    \begin{abstract}
        We consider the semi-supervised clustering problem where crowdsourcing provides noisy information about the pairwise comparisons on a small subset of data, i.e., whether a sample pair is in the same cluster. We propose a new approach that includes a deep generative model (DGM) to characterize low-level features of the data, and a statistical relational model for noisy pairwise annotations on its subset. The two parts share the latent variables. To make the model automatically trade-off between its complexity and fitting data, we also develop its fully Bayesian variant. The challenge of inference is addressed by fast (natural-gradient) stochastic variational inference algorithms, where we effectively combine variational message passing for the relational part and amortized learning of the DGM under a unified framework. Empirical results on synthetic and real-world datasets show that our model outperforms previous crowdsourced clustering methods.
    \end{abstract}
    
    \vspace{-0.4cm}
    \section{Introduction}\label{sec:intro}
    \vspace{-0.1cm}
    
    Clustering is a classic data analysis problem when the taxonomy of data is unknown in advance. Its main goal is to divide samples into disjunct clusters based on the similarity between them. Clustering is useful in various application areas including computer vision~\cite{shi2000normalized}, bioinformatics~\cite{wiwie2015comparing}, anomaly detection~\cite{chandola2009anomaly}, etc.
    When the feature vectors of samples are observed, most clustering algorithms require a similarity or distance metric defined in the feature space, so that the optimization objective can be built.
    Since different metrics may result in entirely different clustering results, and general geometry metrics may not meet the intention of the tasks' designer, many clustering approaches learn the metric from the side-information provided by domain experts~\cite{xing2003distance}, thus the manual labeling procedure of experts could be a bottleneck for the learning pipeline.

    Crowdsourcing is an efficient way to collect human feedbacks~\cite{howe2006rise}. It distributes micro-tasks to a group of ordinal web workers in parallel, so the whole task can be done fast with relatively low cost.
    It has been used on annotating large-scale machine learning datasets such as ImageNet~\cite{deng2009imagenet}, and can also be used to collect side-information for clustering.
    However, directly collecting labels from crowds may lead to low-quality results due to the lack of expertise of workers.
    Consider an example of labeling a set of images of flowers from different species. One could show images to the web workers and ask them to identify the corresponding species, but such tasks require the workers to be experts in identifying the flowers and have all the species in their minds, which is not always possible. A more reasonable and easier task is to ask the workers to compare pairs of flower images and to answer whether they are in the same species or not.
    Then specific clustering methods are required to discover the clusters from the noisy feedbacks.
    
    To solve above clustering problems with pairwise similarity labels between samples from the crowds,
    Crowdclustering~\cite{gomes2011crowdclustering} discovers the clusters within the dataset using a Bayesian hierarchical model. By explicitly modeling the mistakes and preferences of web workers, the outputs will match the human consciousness of the clustering tasks. This method reduces the labeling cost to a great degree compared with expert labeling. However, the cost still grows quadratically as the dataset size grows, so it is still only suitable for small datasets. In this work, we move one step further and consider the \emph{semi-supervised crowdclustering} problem that jointly models the feature vectors and the crowdsourced pairwise labels for only a subset of samples. When we control the size of the subset to be labeled by crowds, the total labeling budget and time can be controlled.
    A similar problem has been discussed by~\cite{yi2012semi}, while the authors use a linear similarity function defined on the low-level object features, and ignore the noise and inter-worker variations in the manual annotations.

    Different from existing approaches, we propose a semi-supervised deep Bayesian model to jointly model the generation of the labels and the raw features for both crowd-labeled and unlabeled samples. Instead of the direct usage of low-level features, we build a flexible deep generative model (DGM) to capture the latent representation of data, which is more suitable to express the semantic similarity than the low-level features. The crowdsourced pairwise labels are modeled by a statistical relational model, and the two parts (i.e., DGM and the relational model) share the same latent variables. We also investigate the fully Bayesian variant of this model so that it can automatically control its complexity.
    Due to the intractability of exact inference, we develop fast (natural-gradient) stochastic variational inference algorithms. To address the challenges in fully Bayesian inference over model parameters, we effectively combine variational message passing and natural gradient updates for the conjugate part (i.e., the relational model and the mixture model) and amortized learning of the nonconjugate part (i.e., DGM) under a unified framework. Empirical results on synthetic and real-world datasets show that our model outperforms previous crowdsourced clustering methods.
    
        \vspace{-0.2cm}
    \section{Semi-crowdsourced deep clustering}\label{sec:semi_cluster}
        \vspace{-0.2cm}
    \begin{wrapfigure}{r}{.4\textwidth}\vspace{-1cm}
        \captionsetup{width=0.39\textwidth}
        \begin{center}
            \resizebox{.35\textwidth}{!}{ 
                \begin{tikzpicture}
                % Define nodes
                \node[obs] (o) {$\bo$};
                \node[latent, above= of o] (x) {$\bx$};
                \node[latent, above= of x] (z) {$\bz$};
                \node[const, left=of o] (gamma) {$\bgamma$};
                \node[obs, left=of z] (L) {$\mathbf{L}$};
                \node[const, left=of L] (alpha) {$\alpha$};
                \node[const, below=of L] (beta) {$\beta$};
                \node[const, right=of z] (pi) {$\bpi$};
                \node[const, right=of x] (musigma) {$\bmu$, $\bSigma$};
                % Connect the nodes
                \edge {z, musigma} {x} ;
                \edge {x,gamma} {o} ;
                \edge {pi} {z};
                \edge {z, alpha, beta} {L};
                % Plates
                \plate {ozx} {(o)(x)(z)} {$N$};
                \plate {worker} {(L) (alpha) (beta)} {$M$};
                
                \end{tikzpicture}}
        \end{center}
        \caption{Semi-crowdsourced Deep Clustering (SCDC).}\vspace{-0.6cm}
        \label{fig:semi-crowd-dgm}
    \end{wrapfigure}
    
    In this section, we propose the semi-crowdsourced clustering with deep generative models for directly modeling the raw data, which enables end-to-end training. We call the model \textit{Semi-crowdsourced Deep Clustering} (SCDC), whose graphical model is shown in \Cref{fig:semi-crowd-dgm}.
    This model is composed of two parts: the raw data model handles the generative process of the observations $\bO$; the crowdsourcing behavior model on labels $\bL$ describes the labeling procedure of the workers.
    The details for each part will be introduced below. 
    \vspace{-0.1cm}
    \subsection{Model the raw data -- deep generative models}
    \vspace{-0.1cm}
    \label{sec:model-dgm}\vspace{-0.1cm}
    We denote the raw data observations by $\bO=\{\bo_1,...,\bo_N\}$. For images, $\bo_n\in \mathbb{R}^D$ denotes the pixel values. For each data point
    $\bo_n$ we have a corresponding latent variable $\bx_n\in \mathbb{R}^d$ and $p(\bo_n|\bx_n,\bgamma)$ is a flexible neural network density model parametrized by $\bgamma$. $p(\bx_n|\bz_n;\bmu,\bSigma)$ is a Gaussian mixture where $\bz_n$ comprises a $1$-of-$K$ binary vector with elements $z_{nk}$ for $k =1,...,K$. Here $K$ denotes the number of clusters. We denote the local latent variables by $\bX = \{\bx_1,...,\bx_N\}$, $\bZ = \{\bz_1,...,\bz_N\}$.
    When real-valued observations are given, the generative process is as follows:
    \begin{align*}
    p(\bZ;\bpi) &= \prod_{n=1}^{N} p(\bz_n;\bpi) = \prod_{n=1}^{N}\prod_{k=1}^{K}\pi_{k}^{z_{nk}},\quad
    p(\bX|\bZ;\bmu,\bSigma) = \prod_{n=1}^{N}\prod_{k=1}^{K} \mathcal{N}(\bx_n; {\bmu}_k, {\bSigma}_k)^{z_{nk}}, \\
    p(\bO|\bX;\bgamma) &= \prod_{n=1}^{N}\mathcal{N}(\bo_n| \mathbf{\bmu}_{\bgamma}(\bx_n), \text{diag}(\bm\sigma^2_{\bgamma}(\bx_n))),
    \end{align*}
    where $\mathbf{\bmu}_{\bgamma}(\cdot)$ and $\bm\sigma^2_{\bgamma}(\cdot)$ are two neural networks parameterized by $\bgamma$. For other types of observations $\bO$, $p(\bo|\bx;\bgamma)$ can be other distributions, e.g. Bernoulli distribution for binary observations. In general, our model is a deep generative model with structured latent variables.
    
    \vspace{-0.1cm}
    \subsection{Model the behavior of each worker -- two-coin Dawid-Skene model}
    \vspace{-0.1cm}
    We collect pairwise annotations provided by $M$ workers. A partially observed $\bL^{(m)}\!\in\!\{0,1, \texttt{NULL}\}^{N_l\times N_l}$ is the annotation matrix of the $m$-th worker, where $N_l$ is the number of annotated data points. For observation pairs $(\bo_i, \bo_j), i\neq j$, $L_{ij}^{(m)}=1$ represents that the $m$-th worker provides a must-link (ML) constraint, which means observations $i$ and $j$ belong to a same cluster, $L_{ij}^{(m)}=0$ represents cannot-link (CL) constraint, which means observations $i$ and $j$ belong to different clusters, and $\texttt{NULL}$ represents that $L_{ij}^{(m)}$ is not observed. 
    It is obvious that $\bL^{(m)}$ is symmetric, i.e., $L_{ij}^{(m)} = L_{ji}^{(m)},\; \forall i, j, m$. Self-edges are not allowed, i.e., $L_{ii}^{(m)} = \texttt{NULL}, \forall i$.
    
    Among all the $N$ data observations $\bO$, 
    we only crowdsource pairwise annotations for a small portion of $\bO$, denoted by $\bO_L$.
    Each worker only provides annotations to a small amount of items in $\bO_L$ and the annotation accuracies of non-expert workers may vary with observations and levels of expertise.
    We adopt the two-coin Dawid-Skene model for annotators from~\cite{raykar2010learning} and develop a probabilistic model by explicitly modeling the uncertainty of each worker.
    Specifically, the uncertainty of the $m$-th worker can be characterized by accuracy parameters $(\alpha_m,\beta_m)$, where $\alpha_m$ represents sensitivity, which means the probability of providing ML constraints for sample pairs belonging to the same cluster. And $\beta_m$ is the $m$-th worker's specificity, which means the probability of providing CL constraints for sample pairs from different clusters. Let $\balpha = \{\alpha_1,...,\alpha_M\}$ and $\bbeta = \{\beta_1,...,\beta_M\}$.
    The likelihood is defined as
    \begin{align}
    &p(L^{(m)}_{ij}|\bz_i,\bz_j; \alpha_m, \beta_m) =\text{Bern}(L^{(m)}_{ij}|\alpha_m)^{\bz_i^\top\bz_j}\text{Bern}(L^{(m)}_{ij}|1\!-\!\beta_m)^{1-\bz_i^\top\bz_j},
    %&=\left[\alpha_m^{L_{ij}^{(m)}}(1 - \alpha_m)^{1 - L_{ij}^{(m)}}\right]^{\bz_i^\top\bz_j}\left[(1 - \beta_m)^{L_{ij}^{(m)}}\beta_m^{1 - L_{ij}^{(m)}}\right]^{1 - \bz_i^\top\bz_j}
    \end{align}
    or equivalently, $p(L^{(m)}_{ij} = 1| \bz_i = \bz_j, \alpha_m) = \alpha_m$, 
    $p(L^{(m)}_{ij} = 0| \bz_i \neq \bz_j, \beta_m) = \beta_m.
    $
    To simplify the notation, we define $I_{ij}^{(m)} = \mathbb{I}[L_{ij}^{(m)} \neq \texttt{NULL}]$. Using the symmetry of $\bL^{(m)}$, the total likelihood of annotations can be written
    \begin{equation}
    \label{equ:ann}
    p(\bL|\bZ;\balpha,\bbeta) = \prod_{m=1}^M\prod_{1\leq i<j\leq N} p(L_{ij}^{(m)}|\bz_i,\bz_j;\alpha_m,\beta_m)^{I_{ij}^{(m)}}.
    \end{equation}

    \vspace{-0.2cm}
    \subsection{Amortized variational inference}
    \vspace{-0.1cm}
    \label{sec:amortized-vi}
    As described above, the parameters in the semi-crowdsourced deep clustering model include $\bpi\in [0, 1]^K$, $\bmu \in \mathbb{R}^{K\times d}$,$\bSigma \in \mathbb{R}^{K\times d\times d}$, $\balpha,\bbeta\in [0, 1]^M$, and the parameters of neural networks $\bgamma$. Let $\bTheta=\{\bpi,\bmu,\bSigma,\balpha,\bbeta\}$, the overall joint likelihood of the model is
    \begin{equation}
    \begin{aligned}
    &p(\bZ,\bX,\bO,\bL;\bTheta,\bgamma)=p(\bZ;\bpi)p(\bX|\bZ;\bmu,\bSigma)p(\bO|\bX;\bgamma)p(\bL|\bZ;\balpha,\bbeta).
    \end{aligned}
    \end{equation}
    %    For those data without annotations $\bO_u$, similarly, we have
    %$p(\bZ,\bX,\bO_u;\bTheta)=p(\bZ;\bpi)p(\bX|\bZ;\bmu,\bSigma)p(\bO_u|\bX;\bgamma).$   
    For this model, the learning objective is to maximize the variational lower bound $\mathcal{L}(\bO, \bL)$ of the marginal log likelihood of the entire dataset $\log p(\bO,\bL)$:
    \begin{equation}
    %    \begin{aligned}
    \log p(\bO,\bL) \geq \mathbb{E}_{q(\bZ, \bX|\bO)}\left[\log p(\bZ,\bX,\bO,\bL;\bTheta,\bgamma) - \log q(\bZ, \bX|\bO)\right] = \mathcal{L}(\bO, \bL; \bTheta,\bgamma,\bphi)\\
    %& \log p(\bO_u)\geq  \mathbb{E}_{q(\bZ, \bX|\bO_u)}\left[\log p(\bZ,\bX,\bO_u;\bTheta) - \log q(\bZ, \bX|\bO_u)\right]=\mathcal{U}(\bO_u) 
    %    \end{aligned}
    \end{equation}
    To deal with the non-conjugate likelihood $p(\bO|\bX;\bgamma)$, we introduce inference networks for each of the latent variables $\bz_n$ and $\bx_n$. The inference networks are assumed to have a factorized form $q(\bz_n,\bx_n|\bo_n) = q(\bz_n|\bo_n;\bphi)q(\bx_n|\bz_n,\bo_n;\bphi)$, which are Categorical and Normal distributions, respectively:
    \begin{equation*}
    %    \hspace{-0.2cm}
    \begin{aligned}
    q(\bz_n|\bo_n;\bphi) = \textrm{Cat}(\bz_n;\bpi(\bo_n;\bphi)),
    q(\bx_n|\bz_n,\bo_n;\bphi) = \mathcal{N}(\bmu(\bz_n,\bo_n;{\bphi}), \mathrm{diag}(\bsigma^2(\bz_n,\bo_n;{\bphi}))),
    \end{aligned}
    \end{equation*}
    where $\bsigma(\bz_n,\bo_n;\bphi)$ is a vector of standard deviations and $\bphi$ denotes the inference networks parameters. Similar to the approach in \cite{kingma2014semi}, we can analytically sum over the discrete variables $\bz_n$ in the lower bound and use the reparameterization trick to compute gradients w.r.t. to $\bTheta,\bgamma$ and $\bphi$.
    
    %    \paragraph{Stochastic approximation} 
    %One key challenge of the gradient optimization is the computational cost, since each iteration requires scanning over all feature vectors ($\bx_{1:N}$) and annotations ($\bL^{(1:M)}$), which is time-consuming for large datasets. 
    The above objective sums over all data and annotations. For large datasets, we can conveniently use a stochastic version by approximating the lower bound with subsampled minibatches of data. 
    %    To address it, we develop a stochastic version of the above variational optimization problem.
    Specifically, the variational lower bound is decomposed into two terms:
    $
    \mathcal{L}(\bL,\bO;\bTheta,\bgamma,\bphi) = \mathcal{L}_{\mathrm{local}} + \mathcal{L}_{\mathrm{rel}}$,
    where
    $
    %    \begin{aligned}
    \mathcal{L}_{\mathrm{local}}
    %= &\mathbb{E}_{q(\bZ,\bX|\bO)}\left[\log p(\bZ,\bX,\bO;\bTheta) - \log q(\bZ|\bO; \bphi) - \log q(\bX|\bZ,\bO;\bphi)\right] \notag \\
    = \sum_{n=1}^N \mathbb{E}_{q(\bz_n,\bx_n|\bo_n)}\left[\log p(\bz_n) + \log p(\bx_n|\bz_n) + \log p(\bo_n|\bx_n) - \log q(\bz_n, \bx_n|\bo_n)\right]$, and $\mathcal{L}_{\mathrm{rel}} =
    \sum_{m=1}^M\sum_{1\leq i<j\leq N} I_{ij}^{(m)} \mathbb{E}_{q(\bz_i,\bz_j|\bo_i,\bo_j;\bphi)} \log p(L_{ij}^{(m)}|\bz_i,\bz_j;\alpha_m,\beta_m).
    %     \end{aligned}
    $
    It is easy to derive an unbiased stochastic approximation of $\mathcal{L}_{\mathrm{local}}$:
    \begin{align*}
    \mathcal{L}_{\mathrm{local}}\approx \frac{N}{|B|}\sum_{n\in B} \mathbb{E}_{q(\bz_n,\bx_n|\bo_n)}\left[\log p(\bz_n) + \log p(\bx_n|\bz_n) + \log p(\bo_n|\bx_n) - \log q(\bz_n, \bx_n|\bo_n)\right],
    \end{align*}
    where $B$ is the sampled minibatch. For $\mathcal{L}_{\mathrm{rel}}$, we can similarly randomly sample a minibatch $S$ of annotations: $
    \mathcal{L}_{\mathrm{rel}}\approx \frac{N_a}{|S|}\sum_{(i, j, m)\in S} \mathbb{E}_{q(\bz_i,\bz_j|\bo_i,\bo_j;\bphi)} \log p(L_{ij}^{(m)}|\bz_i,\bz_j;\alpha_m,\beta_m)$,
    where $N_a = \sum_{m=1}^M\sum_{1\leq i<j\leq N} I_{ij}^{(m)}$ denotes the total number of annotations.
    
    \section{Natural gradient inference for the fully Bayesian model}
    \vspace{-0.1cm}
    
    \label{sec:bayes}

    In the previous section, the global parameters $\bTheta = \{\balpha, \bbeta, \bpi, \bmu, \bSigma\}$
    %   \footnote{We redefined the notation $\bTheta$ in \Cref{sec:amortized-vi}.} 
    are assumed to be deterministic and are directly optimized by gradient descent. 
    In this section, we propose a fully Bayesian variant of our model (BayesSCDC), which has an automatic trade-off between its complexity and fitting the data. There is no overfitting if we choose a large number $K$ of components in the mixture, in which case the variational treatment below can automatically determine the optimal number of mixture components. We develop fast natural-gradient stochastic variational inference algorithms for BayesSCDC, which effectively combines variational message passing for the conjugate structures (i.e., the relational part and the mixture part) and amortized learning of deep components (i.e., the deep generative model).

    \vspace{-0.1cm}
    \subsection{Fully Bayesian semi-crowdsourced deep clustering (BayesSCDC)}
    \vspace{-0.1cm}
    For the mixture model, we choose a Dirichlet prior over the mixing coefficients $\bpi$ and an independent Normal-Inverse-Wishart prior governing the mean and covariance $(\bmu,\bSigma)$ of each Gaussian component, given by
    \begin{equation}
    p(\bpi) = \text{Dir}(\bpi|\alpha_0) = C(\alpha_0)\prod_{k=1}^{K} \pi_{k}^{\alpha_0-1},\quad
    p(\bmu,\bSigma) = \prod_{k=1}^{K}\text{NIW}(\bmu_k,\bSigma_k|\bbm,\kappa, \bS,\nu),
    \end{equation}
    where $\bbm\in \mathbb{R}^d$ is the location parameter, $\kappa > 0$ is the concentration, $\bS\in\mathbb{R}^{d\times d}$ is the scale matrix (positive definite), and $\nu> d - 1$ is the degrees of freedom.
    The densities of $\bpi, \bz, (\bmu,\bSigma),\bx$ can be written in the standard form of exponential families as:
    \begin{align*}
    &p(\bpi)
    =
    \exp \left\{ \langle \bmeta_\bpi^0, \, \bt(\bpi) \rangle - \log Z(\bmeta_\bpi^0) \right\},\quad p(\bmu,\bSigma)
    =
    \exp \left\{ \langle \bmeta_{\bmu,\bSigma}^0,\, \bt(\bmu,\bSigma) \rangle\! -\! \log Z(\bmeta_{\bmu,\bSigma}^0) \right\},\\
    &p(\bz|\bpi)
    =
    \exp \left\{ \langle \bmeta_\bz^0(\bpi), \, \bt(\bz) \rangle - \log Z(\bmeta_\bz^0(\bpi)) \right\} = \exp\left\{\langle\bt(\bpi), (\bt(\bz), \mathbf{1})\rangle\right\}, \\
    &p(\bx|\bz,\bmu,\bSigma)
    =
    \exp \left\{ \langle \bt(\bz), \, \bt(\bmu,\bSigma)^\top(\bt(\bx), \mathbf{1}) \rangle  \right\},
    \end{align*}
    where $\bmeta$ denotes the natural parameters, $\bt(\cdot)$ denotes the sufficient statistics\footnote{Detailed expressions of each distribution can be found in \Cref{app:deriv}}, and $\log Z(\cdot)$ denotes the log partition function.
    
    For the relational model, we assume the accuracy parameters of all workers ($\balpha$, $\bbeta$) are drawn independently from common priors. We choose conjugate Beta priors for them as
    \begin{align}
    &p(\balpha) = \prod_{m=1}^{M}p(\alpha_m) = \prod_{m=1}^{M}\text{Beta}(\tau_{\alpha_0^1}, \tau_{\alpha_0^2}),\quad
    p(\bbeta) = \prod_{m=1}^{M}p(\beta_m) = \prod_{m=1}^{M}\text{Beta}(\tau_{\beta_0^1}, \tau_{\beta_0^2}).
    \end{align}
    We write the exponential family form of $p(\alpha_m)$ as: $p(\alpha_m)\!=\!\exp \left\{\langle\bmeta_{\alpha_m}^0, \bt(\alpha_m)\rangle\!-\!\log Z(\bmeta_{\alpha_m}^0)\right\}$ ($p(\beta_m)$ is similar),
    where $\bmeta_{\alpha_m}^0 = [\tau_{\alpha_0^1}-1, \tau_{\alpha_0^2}-1]^\top$ and $ \bt(\alpha_m) = \left[\log \alpha_m, \log (1- \alpha_m)\right]^\top$.
    \vspace{-0.1cm}
    \subsection{Natural-gradient stochastic variational inference}
    \vspace{-0.1cm}
    \label{sec:natural}
    %\paragraph{Natural-gradient stochastic variational inference}
    The overall joint distribution of all of the hidden and observed variables takes the form:
    \begin{align}
    p(\bL^{(1:M)}, \bO,\bX,\bZ,\bTheta;\bgamma) = & ~p(\bpi)p(\bZ|\bpi)p(\bmu,\bSigma)p(\bX|\bZ,\bmu,\bSigma)p(\bO|\bX;\bgamma) \\ \notag
    & ~\cdot p(\balpha)p(\bbeta)p(\bL^{(1:M)}|\bZ,\balpha,\bbeta).
    \end{align}
    Our learning objective is to maximize the marginal likelihood of observed data and pairwise annotations $\log p(\bO, \bL^{(1:M)})$.
    Exact posterior inference for this model is intractable.
    Thus we consider a mean-field variational family
    $q(\bTheta,\bZ,\bX) = q(\balpha)q(\bbeta)q(\bZ)q(\bX)q(\bpi)q(\bmu,\bSigma).$
    To simplify the notations, 
    we write each variational distribution in its exponential family form: $
    q(\btheta) =
    \exp \left\{ \langle \bmeta_{\btheta}, \, \bt(\btheta) \rangle - \log Z(\bmeta_\btheta) \right\},\; \btheta \in \bTheta \cup \bZ \cup \bX.$
    The evidence lower bound (ELBO) ${\mathcal{L}}(\bmeta_{\bTheta},\bmeta_{\bZ},\bmeta_{\bX};\bgamma)$ of $\log p(\bO, \bL^{(1:M)})$ is 
    \begin{align} \label{eq:fully-bayesian-elbo}
    &\log p(\bO,\bL^{(1:M)}) \geq {\mathcal{L}}(\bmeta_{\bTheta},\bmeta_{\bZ},\bmeta_{\bX};\bgamma) \triangleq\mathbb{E}_{q(\bTheta, \bZ,\bX)}\log\left[\frac{p(\bL^{(1:M)}, \bO,\bX,\bZ,\bTheta;\bgamma)}{q(\bTheta)q(\bZ)q(\bX)}\right].
    \end{align}
    In traditional mean-field variational inference for conjugate models, the optimal solution of maximizing \cref{eq:fully-bayesian-elbo} over each variational parameter can be derived analytically given other parameters fixed, thus a coordinate ascent can be applied as an efficient message passing algorithm \cite{winn2005variational,hoffman2013stochastic}. However,  it is not directly applicable to our model due to the non-conjugate observation likelihood $p(\bO|\bX;\bgamma)$.  Inspired by \cite{johnson2016composing}, we handle the non-conjugate likelihood by introducing recognition networks $r(\bo_i;\bphi)$. Different from SCDC in \Cref{sec:amortized-vi}, the recognition networks here are used to form conjugate graphical model potentials:
    %     rather than the complete variational distribution's parameters as in VAE~\cite{kingma2013auto}: %\tian{Explain the benefits here.}
    \begin{equation}
    \label{eq:recognition}
    \psi(\bx_i;\bo_i, \bphi)\triangleq \langle r(\bo_i;\bphi), \bt(\bx_i)\rangle.
    \end{equation}
    By replacing the non-conjugate likelihood $p(\bO|\bX;\bgamma)$ in the original ELBO with a conjugate term defined by $\psi(\bx_i;\bo_i, \bphi)$, we have the following surrogate objective $\hat{\mathcal{L}}$:
    \begin{align} \label{eq:surrogate-elbo}
    &\widehat{\mathcal{L}} (\bmeta_{\bTheta},\bmeta_{\bZ},\bmeta_{\bX};\bphi) \triangleq \mathbb{E}_{q(\bTheta, \bZ,\bX)}\!\log\!\left[\frac{p(\bL^{(1:M)},\bX,\bZ,\bTheta)\exp\{\psi(\bX;\bO, \bphi)\}}{q(\bTheta)q(\bZ)q(\bX)}\!\right].
    \end{align}
    As we shall see, the surrogate objective $\widehat{\mathcal{L}}$ helps us exploit the conjugate structure in the model, thus enables a fast message-passing algorithm for these parts. 
    %    while keeping the benefits of a neural-network observation likelihood. 
    Specifically, we can view \cref{eq:surrogate-elbo} as the ELBO of a conjugate graphical model with the same structure as in Fig.~\ref{fig:semi-crowd-dgm} (up to a constant). Similar to coordinate-ascent mean-field variational inference~\cite{hoffman2013stochastic}, we can derive the local partial optimizers of individual variational parameters as below.  
    
    The optimal solution for $\opt q(\bX)$ factorizes over $n$ , i.e., $\opt q(\bX) = \prod_{i=1}^{N} \opt q(\bx_i)$, and $\opt q(\bx_i)$ depends on the expected sufficient statistics of $(\bmu,\bSigma)$ and $\bz_n$:
    \begin{align}
    \log \opt q(\bx_i)&=\mathbb{E}_{q(\bmu,\bSigma)q(\bz_i)} \log p(\bx_i|\bz_i,\bmu,\bSigma)
    +\langle r(\bo_i;\bphi), \bt(\bx_i)\rangle+\text{const},\\
    \bmeta_{\bx_i}^*&=\mathbb{E}_{q(\bmu,\bSigma)}[\bmeta_{\bx_i}^0(\bmu,\bSigma)]^\top\mathbb{E}_{q(\bz_i)}[\bt(\bz_i)]\!+\!r(\bo_i;\bphi).
    \label{eq:etax}
    \end{align}
    %    However, $\opt q(\bZ)$ cannot factorize into independent terms of $\opt q(\bz_i)$ because the Markov blanket of $\bz_i$ includes $\bZ_{\mathcal{I}_i}$ as a result of $\bL^{(1:M)}$,
    By further assuming a mean-field structure over $\bZ$: $\opt q(\bZ) = \prod_{i=1}^N \opt q(\bz_i)$, we have the local partial optimizer for each single $q(\bz_i)$ as
    \begin{align}
    &\log \opt q(\bz_i) = \mathbb{E}_{q(\bpi)}\log p(\bz_i|\bpi)+ \mathbb{E}_{q(\bmu,\bSigma)q(\bx_i)}\log p(\bx_i|\bz_i,\bmu,\bSigma)\notag\\
    &\quad\quad\quad\quad\quad+\mathbb{E}_{q(\balpha)q(\bbeta)q(\bZ_{-i})}\left[\log p(\bL^{(1:M)}|\bZ,\balpha,\bbeta)\right] + \text{const}, \\
    &\bmeta_{\bz_i}^* =  \mathbb{E}_{q(\bpi)}\bt(\bpi) + \mathbb{E}_{q(\bmu,\bSigma)}\left[\bt(\bmu,\bSigma)\right]^\top\mathbb{E}_{q(\bx_i)}\left[(\bt(\bx_i), \mathbf{1})\right] + \sum_{m\!=\!1}^M\sum_{j=1}^N w_{ij}^{(m)}\mathbb{E}_{q(\bz_j)}[\bt(\bz_j)],
    \label{eq:etaz}
    \end{align}
    where $w_{ij}^{(m)} = I_{ij}^{(m)}\mathbb{E}_{q(\balpha, \bbeta)}\left[\ln\frac{1-\alpha_m}{\beta_m}+L_{ij}^{(m)}\left(\ln\frac{\alpha_m}{1-\alpha_m} + \ln\frac{\beta_m}{1-\beta_m}\right)\right]$ is the weight of the message from $\bz_j$ to $\bz_i$.
    Using a block coordinate ascent algorithm that applies \cref{eq:etax,eq:etaz} alternatively, 
    we can find the joint local partial optimizers $\left(\bmeta_\bZ^*(\bmeta_\bTheta, \bphi), \bmeta_\bX^*(\bmeta_\bTheta, \bphi)\right)$ of $\widehat{\mathcal{L}}$ w.r.t. $\left(\bmeta_\bX, \bmeta_\bZ\right)$ given other parameters fixed, i.e., 
    \begin{eqnarray}
    \nabla_{\bmeta_\bZ}\widehat{\mathcal{L}}(\bmeta_{\bTheta},\bmeta_{\bZ}^*(\bmeta_\bTheta, \bphi),\bmeta_{\bX}^*(\bmeta_\bTheta, \bphi),\bphi) = 0,\quad
    \nabla_{\bmeta_\bX}\widehat{\mathcal{L}}(\bmeta_{\bTheta},\bmeta_{\bZ}^*(\bmeta_\bTheta, \bphi),\bmeta_{\bX}^*(\bmeta_\bTheta, \bphi),\bphi) = 0.
    \end{eqnarray}
    Plugging $\left(\bmeta_\bZ^*(\bmeta_\bTheta, \bphi), \bmeta_\bX^*(\bmeta_\bTheta, \bphi)\right)$ back into $\mathcal{L}$, we define the final objective
    \begin{equation} \label{eq:svae}
    \mathcal{J} (\bmeta_{\bTheta};\bphi,\bgamma)\triangleq \mathcal{L}(\bmeta_{\bTheta},\bmeta_{\bZ}^*(\bmeta_{\bTheta},\bphi), \bmeta_{\bX}^*(\bmeta_{\bTheta},\bphi),\bgamma).
    \end{equation}
    As shown in \cite{johnson2016composing}, $\mathcal{J}(\bmeta_{\bTheta};\bphi, \bgamma)$ lower-bounds the partially-optimized mean field objective, i.e., 
    $
    %    \max_{q(\bX,\bZ)}\mathcal{L}[q(\bTheta,\bZ,\bX),\bgamma] \geq 
    \max_{\bmeta_\bX,\bmeta_\bZ} \mathcal{L}(\bmeta_\bTheta,\bmeta_\bZ,\bmeta_\bX, \bgamma)\geq \mathcal{J}(\bmeta_\bTheta,\bgamma,\bphi),
    $
    %    The KL divergence terms in \cref{eq:svae} have simple forms because they are between members of the same tractable exponential families.
    thus can serve as a variational objective itself.
    We compute the natural gradients of $\mathcal{J}$ w.r.t. the global variational parameters $\bmeta_\bTheta$:
    \begin{align}
    \label{eq:grad_theta}
    \widetilde{\nabla}_{\bmeta_{\bTheta}}\mathcal{J}=&~ \left[\bmeta_\bTheta^0 + \mathbb{E}_{\opt q(\bZ)\opt q(\bX)}\left(\bt(\bZ,\bX,\bL^{(1:M)}),\mathbf{1}\right) -  \bmeta_\bTheta\right] \notag\\
    &~+ \left(\nabla_{\bmeta_\bZ, \bmeta_\bX} \mathcal{L}(\bmeta_{\bTheta},\bmeta_{\bZ}^*(\bmeta_\bTheta, \bphi),\bmeta_{\bX}^*(\bmeta_\bTheta, \bphi); \bgamma), \mathbf{0}\right).
    \end{align}
    Note that the first term in \cref{eq:grad_theta} is the same as the formula of natural gradient in SVI~\cite{hoffman2013stochastic}, which is easy to compute, and the second term originates from the dependence of $\bmeta_{\bZ}^*, \bmeta_{\bX}^*$ on $\bmeta_\bTheta$ and can be computed using the reparameterization trick.
    For other parameters $\bphi, \bgamma$, we can also get the gradients $\nabla_\bphi
    \mathcal{J}(\bmeta_\bTheta;  \bphi,\bgamma)$ and $\nabla_{\bgamma} \mathcal{J}(\bmeta_\bTheta; \bphi, \bgamma)$ using the reparameterization trick.
    \paragraph{Stochastic approximation:} Computing the full natural gradient in \cref{eq:grad_theta} requires to scan over all data and annotations, which is time-consuming. Similar to \Cref{sec:amortized-vi}, we can approximate the variational lower bound with unbiased estimates using mini-batches of data and annotations, thus getting a stochastic natural gradient. Several sampling strategies have been developed for relational model~\cite{gopalan2012scalable} to keep the stochastic gradient unbiased. Here we choose the simplest way: we sample annotated data pairs uniformly from the annotations and form a subsample of the relational model, and do local message passing (\cref{eq:etax,eq:etaz}), then perform the global update using stochastic natural gradient calculated in the subsample. Besides, for all the unannotated data, we also subsample mini-batches from them and perform local and global steps without relational terms. The algorithm of BayesSCDC is shown in~\Cref{alg:example}.

    \paragraph{Comparison with SCDC} BayesSCDC is different in two aspects: (a) fully Bayesian treatment of global parameters; (b) variational algorithms. As we shall see in experiments, the result of (a) is that BayesSCDC can automatically determine the number of mixture components during training.
    As for (b), note that the variational family used in SCDC is not more flexible, but more restricted compared to BayesSCDC. In BayesSCDC, the mean-field $q(\bz)q(\bx)$ doesn't imply that $q^*(\bz)$ and $q^*(\bx)$ are independent, instead they implicitly influence each other through message passing in Eqs. (12) and (14). More importantly, in BayesSCDC the variational posterior gathers information from $\bL$ through message passing in the relational model. In contrast, the amortized form $q(\bz|\bo)q(\bx,\bz|\bo)$ used in SCDC ignores the effect of observed annotations $\bL$. Another advantage of the inference algorithm in BayesSCDC is in the computational cost. As we have seen in~\Cref{alg:example}, the number of passes through the $\bx$ to $\bo$ network is no longer linear with $K$ because we get rid of summing over $\bz$ in the observation term as in \Cref{sec:amortized-vi}.
    
    \begin{algorithm}[tb]
        \caption{Semi-crowdsoursed clustering with DGMs (BayesSCDC)}
        \label{alg:example}
        \begin{algorithmic}
            \STATE {\bfseries Input:} observations $\bO = \{\bo_1,...,\bo_N\}$, annotations $\bL^{(1:M)}$, variational parameters $(\bmeta_\bTheta, \bgamma,\bphi)$
            \REPEAT
            \STATE $\psi_i \gets \langle r(\bo_i; \bphi), \bt(\bx_i)\rangle, i=1,...,N$
            \FOR{each local variational parameter $\bmeta_{\bx_i}^*$ and $\bmeta_{\bz_i}^*$}
            \STATE Update alternatively using~\cref{eq:etax} and~\cref{eq:etaz}
            \ENDFOR
            \STATE Sample $\hat \bx_i \sim \opt q(\bx_i), i=1,...,N$
            %    \STATE Sample $\hat\bgamma\sim q(\bgamma)$
            \STATE Use $\hat \bx_i$ to approximate $\mathbb{E}_{\opt q(\bx)}\log p(\bo|\bx;\bgamma)$ in the lower bound $\mathcal{J}$~\cref{eq:svae} 
            \STATE Update the global variational parameters $\bmeta_\bTheta$ using the natural gradient in \cref{eq:grad_theta} %$\mathcal{L}\gets\log p(\bo_i|\hat\bx_i,\hat\bgamma) + $ 
            \STATE Update $\bphi,\bgamma$ using $\nabla_{\bphi,\bgamma}
            \mathcal{J}(\bmeta_\bTheta; \bphi,\bgamma)$	
            \UNTIL{Convergence}
        \end{algorithmic}
    \end{algorithm}%\vspace{-0.3cm}
    
    \vspace{-0.2cm}
    \section{Related work}
    \vspace{-0.2cm}
    
    Most previous works on learning-from-crowds are about aggregating noisy crowdsourced labels from several predefined classes~\cite{dawid1979maximum,raykar2010learning,welinder2010multidimensional,zhouaggregating,Tian2015Max}.
    A common way they use is to simultaneously estimate the workers' behavior models and the ground truths.
    Different from this line of work, crowdclustering~\cite{gomes2011crowdclustering} collects pairwise labels, including the must-links and the cannot-links, from the crowds, then discovers the items' affiliations as well as the category structure from these noisy labels, so it can be used on a border range of applications compared with the classification methods.
    Recent work~\cite{Vinayak2016Crowdsourced} also developed crowdclustering algorithm on triplet annotations.
    
    One shortcoming of crowdclustering is that it can only cluster objects with available manual annotations. 
    For large-scale problems, it is not feasible to have each object manually annotated by multiple workers. 
    Similar problems were extensively discussed in the semi-supervised clustering area, where we are given the features for all the items and constraints on only a same portion of the items.
    Metric learning methods, including Information-Theoretic Metric Learning (\textbf{ITML})~\cite{davis2007information} and Metric Pairwise Constrained KMeans (\textbf{MPCKMeans})~\cite{bilenko2004integrating}, are used on this problem, they first learn the similarity metric between items mainly based on the supervised portion of data, then cluster the rest items using this metric.
    Semi-crowdsourced clustering~(\textbf{SemiCrowd})~\cite{yi2012semi} combines the idea of crowdclustering and semi-supervised clustering, it aims to learn a pairwise similarity measure from the crowdsourced labels of $n$ objects ($n\ll N$) and the features of $N$ objects. Unlike crowdclustering, the number of clusters in SemiCrowd is assumed to be given a priori. And it doesn't estimate the behavior of different workers. Multiple Clustering Views from the Crowd (\textbf{MCVC})~\cite{chang2017multiple} extends the idea to discover several different clustering results from the noisy labels provided by uncertain experts.
    A common shortcoming of these semi-crowdsourced clustering methods is they cannot make good use of unlabeled items when measuring the similarities, while our model is a step towards this direction. 
    
    As shown in \Cref{sec:model-dgm}, our model is a deep generative model (DGM) with relational latent structures. DGMs are a kind of probabilistic graphical models that use neural networks to parameterize the conditional distribution between random variables. Unlike traditional probabilistic models, DGMs can directly model high-dimensional outputs with complex structures, which enables end-to-end training on real data. They have shown success in image generation~\cite{kingma2013auto}, semi-supervised learning~\cite{kingma2014semi}, and one-shot classification~\cite{rezende2016one}. Typical inference algorithms for DGMs are in the amortized form like that in \Cref{sec:amortized-vi}. However, this approach cannot leverage the conjugate structures in latent variables. Therefore few works have been done on fully Bayesian treatment of global parameters in DGMs. \cite{johnson2016composing,lin2018variational} are two exceptions. In \cite{johnson2016composing} the authors propose using recognition networks to produce conjugate graphical model potentials, so that traditional variational message passing algorithms and natural gradient updates can be easily combined with amortized learning of network parameters. Our work extends their algorithm to relational observations, which has not been investigated before.
    
    \vspace{-0.2cm}
    \section{Experiments}
    \vspace{-0.1cm}
    In this section, we demonstrate the effectiveness of the proposed methods on synthetic and real-world datasets with simulated or crowdsourced noisy annotations. Code is available at \url{https://github.com/xinmei9322/semicrowd}. Part of the implementation is based on ZhuSuan~\citep{zhusuan2017}.
    
    \vspace{-0.2cm}
    \subsection{Toy Pinwheel dataset}
    \vspace{-0.1cm}
    
    \textbf{Simulating noisy annotations from workers.}  Suppose we have $M$ workers with accuracy parameters $(\alpha_m,\beta_m)$. We random sample pairs of items $\bo_i$ and $\bo_j$ and generate the annotations provided by worker $m$ based on the true clustering labels of $\bo_i$ and $\bo_j$ as well as the worker's accuracy parameters $(\alpha_m,\beta_m)$. If $\bo_i$ and $\bo_j$ belong to the same cluster, the worker has probability $\alpha_m$ to provide ML constraint $L_{ij}^{(m)} = 1$. If not, the worker has probability $\beta_m$ to provide CL constraint $L_{ij}^{(m)} = 0$.
    
    \textbf{Evaluation metrics.}  The clustering performance is evaluated by the commonly used normalized mutual information (NMI) score~\cite{strehl2002cluster}, measuring the similarity between two partitions. Following recent work~\cite{xie2016unsupervised}, we also report the unsupervised clustering accuracy, which requires to compute the best mapping using the Hungarian algorithm efficiently.

    \begin{figure}[t]
        \centering
        \begin{subfigure}[t]{.23\textwidth}
            \centering
            \includegraphics[height=3cm]{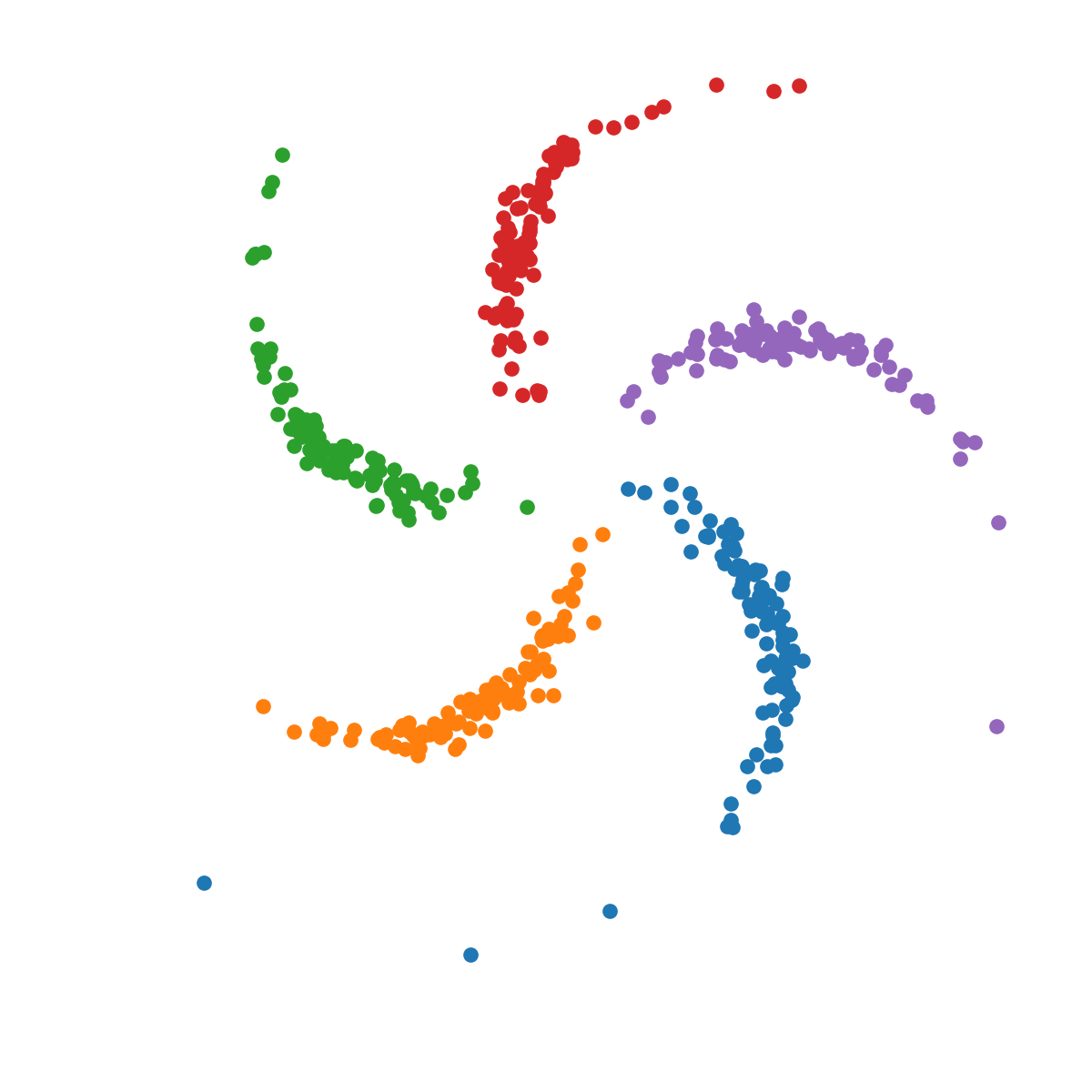} \vspace{-.1cm}
            \caption{The Pinwheel dataset.}
            \label{fig:data}
        \end{subfigure}
        \hfill
        \begin{subfigure}[t]{.23\textwidth}
            \centering
            \includegraphics[height=3cm]{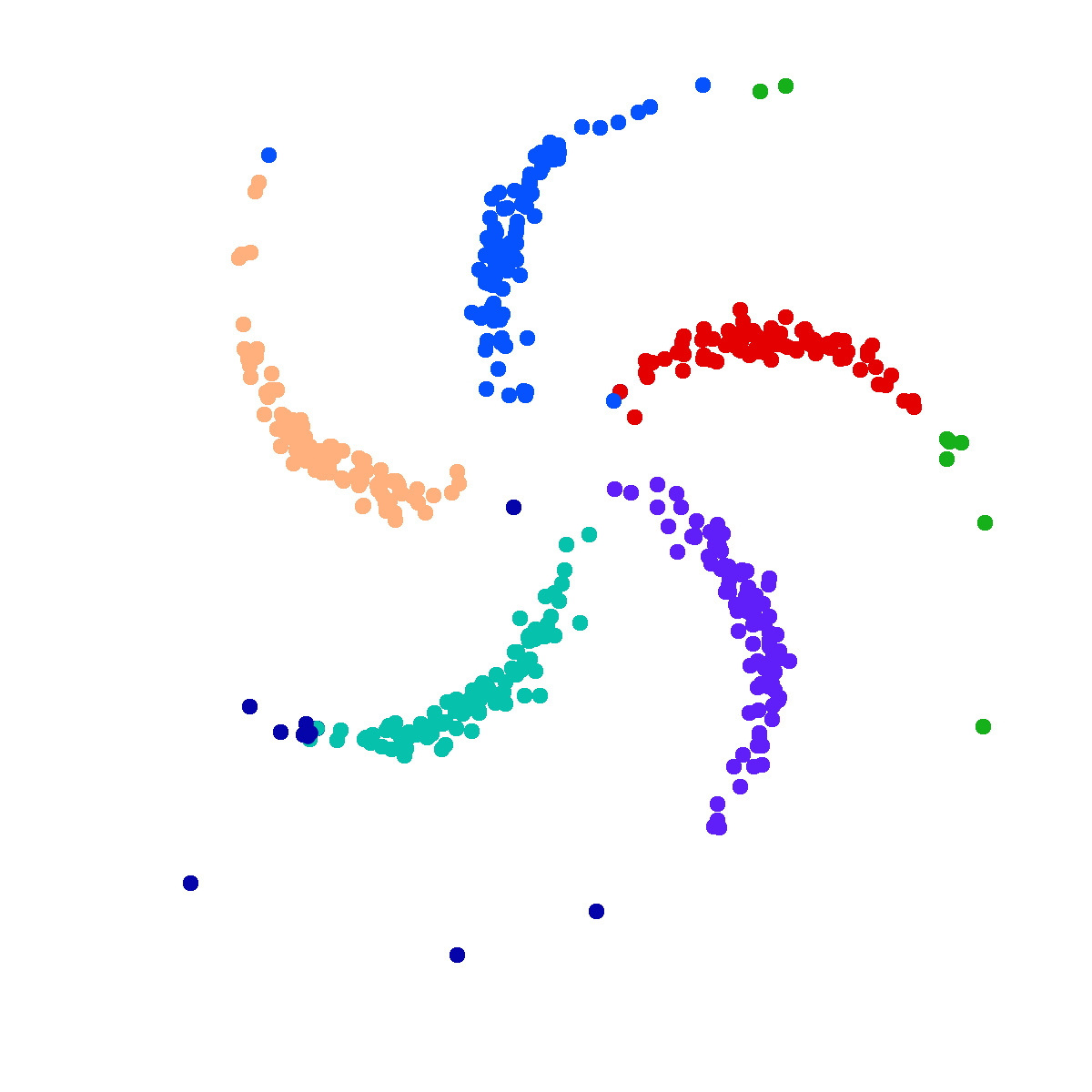} \vspace{-.1cm}
            \caption{Without annotations, good initialization.}
            \label{fig:svae_good}
        \end{subfigure}
        \hfill
        \begin{subfigure}[t]{.23\textwidth}
            \centering
            \includegraphics[height=3cm]{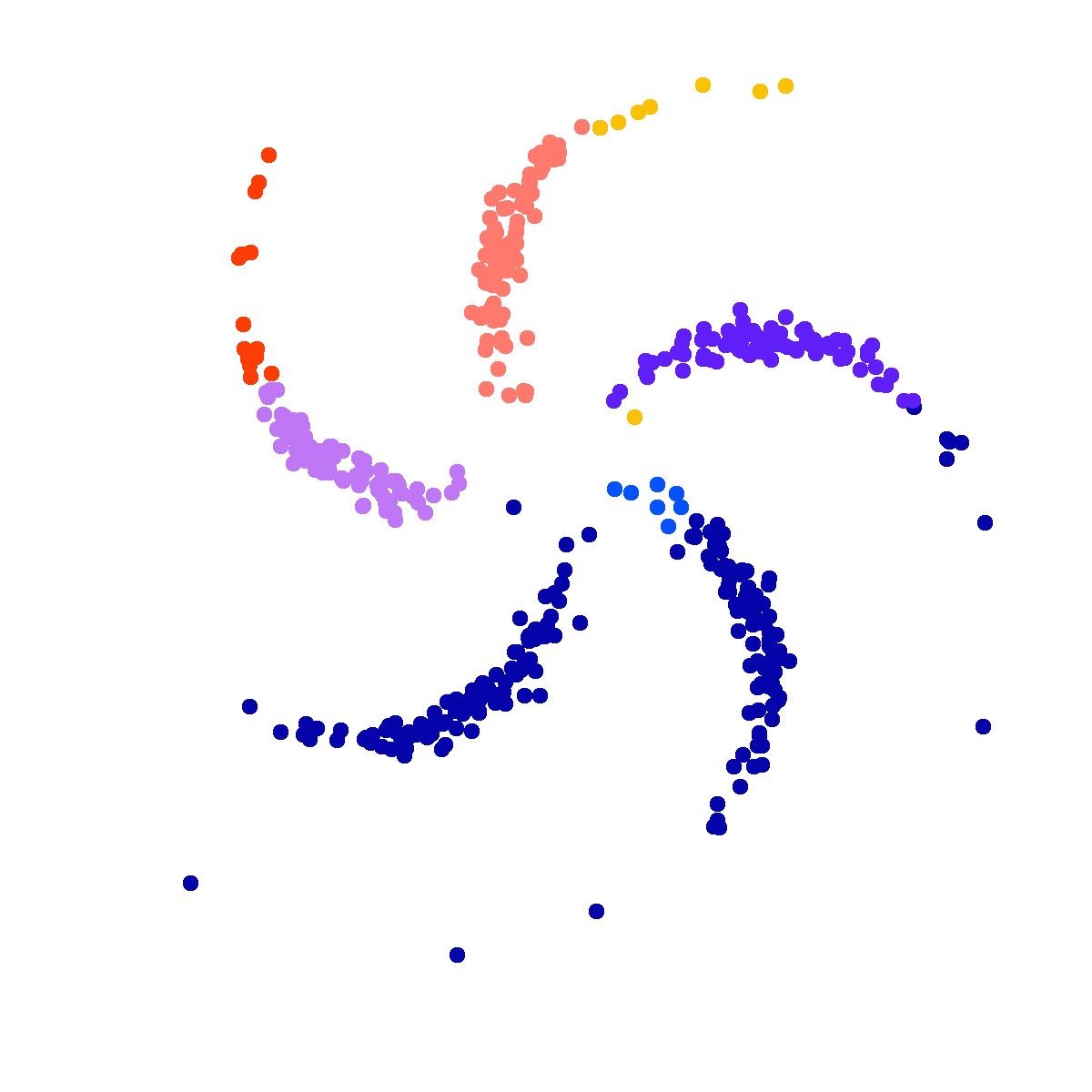}\vspace{-.1cm}
            \caption{Without annotations, bad initialization.}
            \label{fig:svae_bad}
        \end{subfigure}
        \hfill
        \begin{subfigure}[t]{.23\textwidth}
            \centering
            \includegraphics[height=3cm]{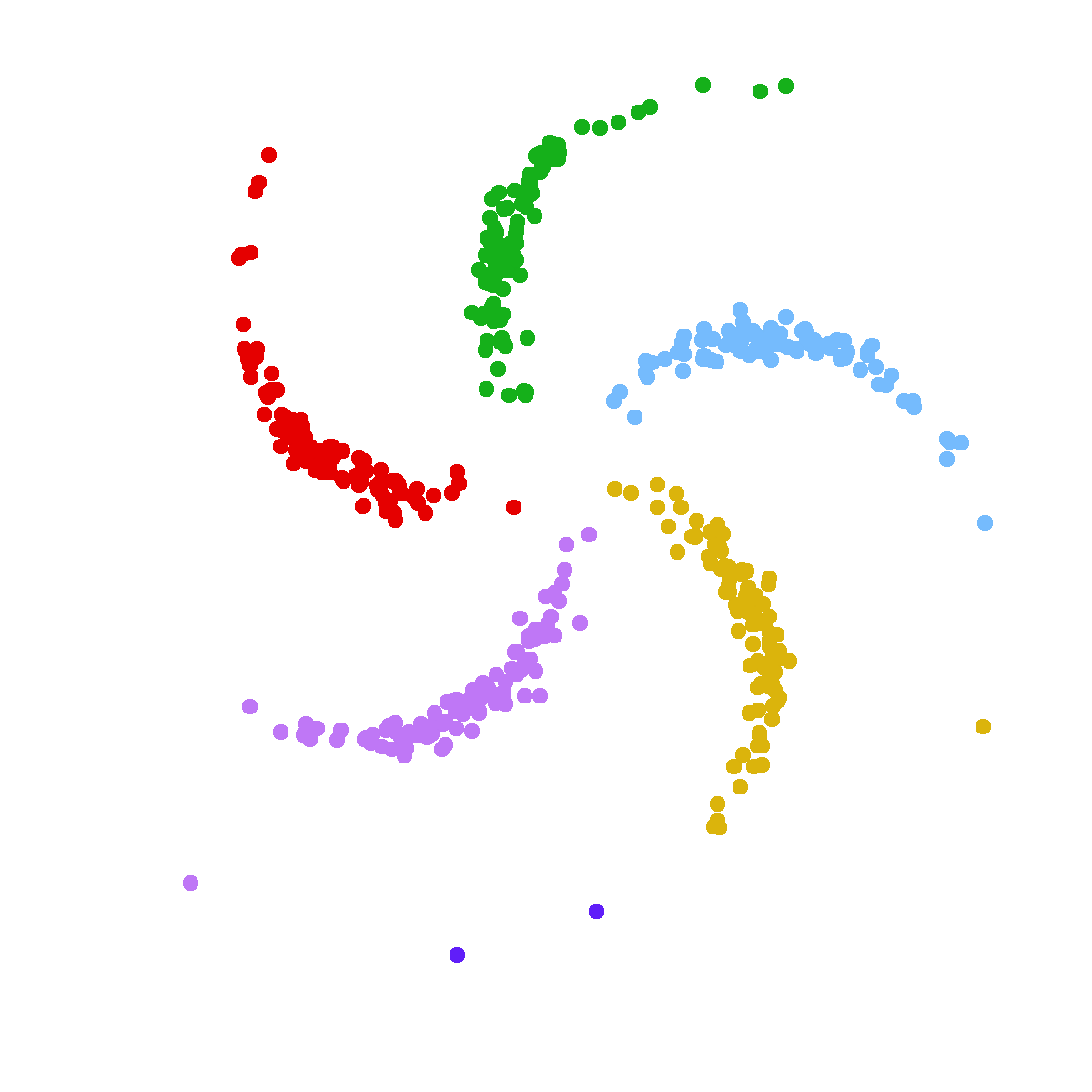}\vspace{-.1cm}
            \caption{With noisy annotations on a subset of data.}
            \label{fig:scc_good}
        \end{subfigure}
        \caption{Clustering results on the Pinwheel dataset, with each color representing one cluster.}
        \label{fig:synthetic}\vspace{-0.2cm}
    \end{figure}
    
    First we apply our method to a toy example--the pinwheel dataset in Fig.~\ref{fig:data} following~\cite{johnson2016composing,lin2018variational}. It has 5 clusters and each cluster has 100 data points, thus there are 500 data points in total. We compare with unsupervised clustering to understand the help of noisy annotations. The clustering results are shown in Fig.~\ref{fig:synthetic}.
    We random sampled 100 data points for annotations and simulate 20 workers, each worker gives 49 pairs of annotations, 980 in total. We set equal accuracy to each worker $\alpha_m=\beta_m=0.9$. 
    
    We use the fully Bayesian model (BayesSCDC) described in \Cref{sec:bayes}. The initial number of clusters is set to a larger number $K=15$ since the hyper priors have sparsity property intrinsically and can learn the number of clusters automatically.
    Unsupervised clustering is sensitive to the initializations, which achieves 95.6\% accuracy and NMI score 0.91 with good initializations as shown in Fig.~\ref{fig:svae_good}. After training, it learns $K=8$ clusters. However, with bad initializations, the accuracy and NMI score of unsupervised clustering are 75.6\% and 0.806, respectively, as shown in Fig.~\ref{fig:svae_bad}.
    With noisy annotations on random sampled 100 data points, our model improves accuracy to 96.6\% and NMI score to 0.94. And it converges to $K=6$ clusters. Our model prevents the bad results in Fig.~\ref{fig:svae_bad} by making use of annotations.

    \vspace{-0.05cm}
    \subsection{UCI benchmark experiments}
    \label{sec:uci}
    In this subsection, we compare the proposed SCDC with the competing methods on the UCI benchmarks. The baselines include \textbf{MCVC}~\cite{chang2017multiple}, \textbf{SemiCrowd}~\cite{yi2012semi}, semi-supervised clustering methods such as \textbf{ITML}~\cite{davis2007information}, \textbf{MPCKMeans}~\cite{bilenko2004integrating} and Cluster-based Similarity Partitioning Algorithm (\textbf{CSPA})~\cite{strehl2002cluster}. 
    
    Crowdsourced annotations are not available for UCI datasets. Following the experimental protocol in \textbf{MCVC}~\cite{chang2017multiple}, we generate noisy annotations given by $M=5$ simulated workers with different sensitivity and specificity, i.e., $\balpha=\bbeta=[0.95,0.9,0.85, 0.8,0.75]$, which is more challenging than equal accuracy parameters. The annotations provided by each worker varies from $200$ to $2000$ and the number of ML constraints equals to the number of CL constraints.
    
\begin{figure}[t]
        \centering
        \begin{subfigure}[t]{.32\textwidth}
            \centering
            \includegraphics[height=3.3cm]{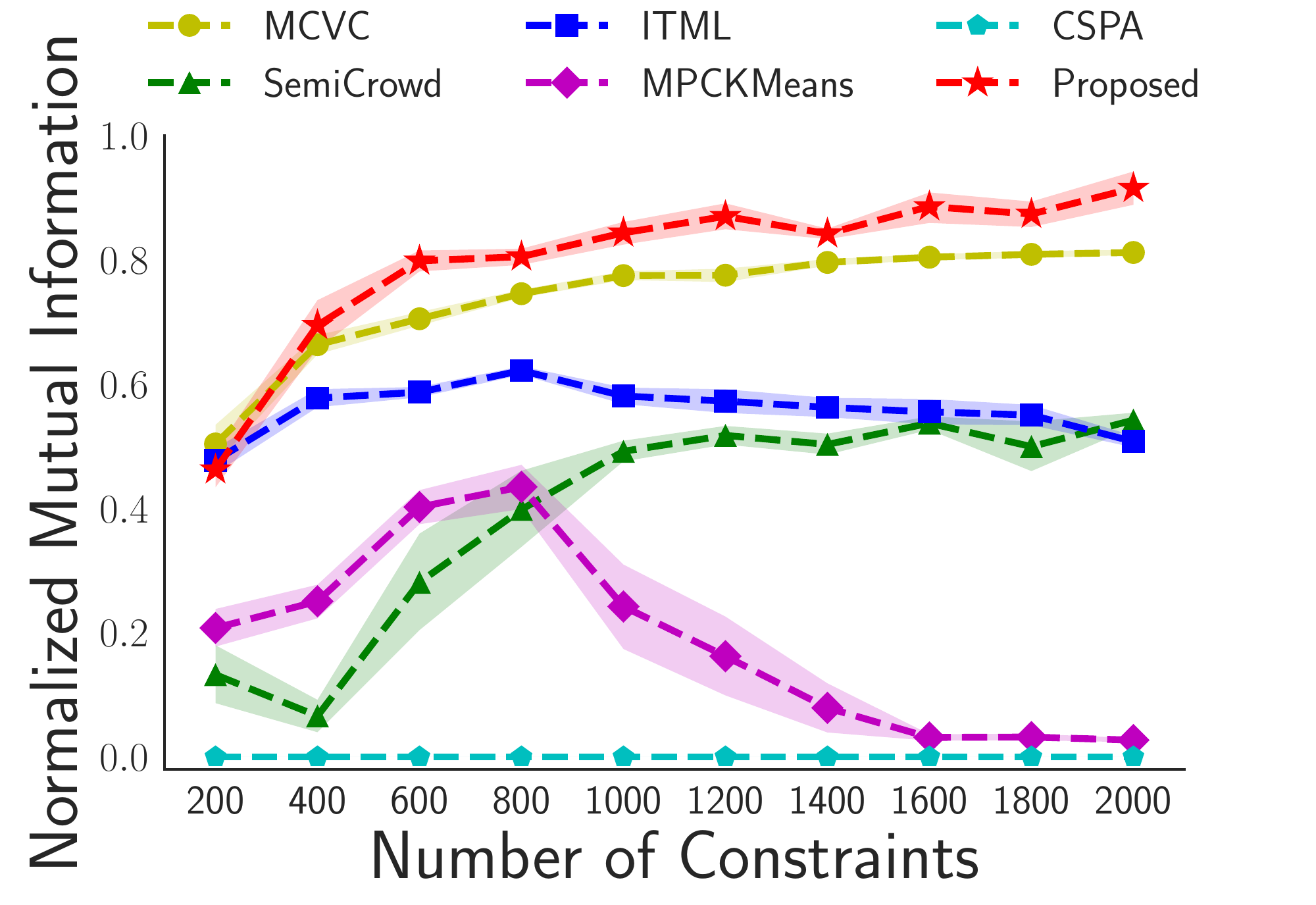}
            \caption{}
            \label{fig:face640}
        \end{subfigure}
        \hfill
        \begin{subfigure}[t]{.32\textwidth}
            \centering
            \includegraphics[height=3.3cm]{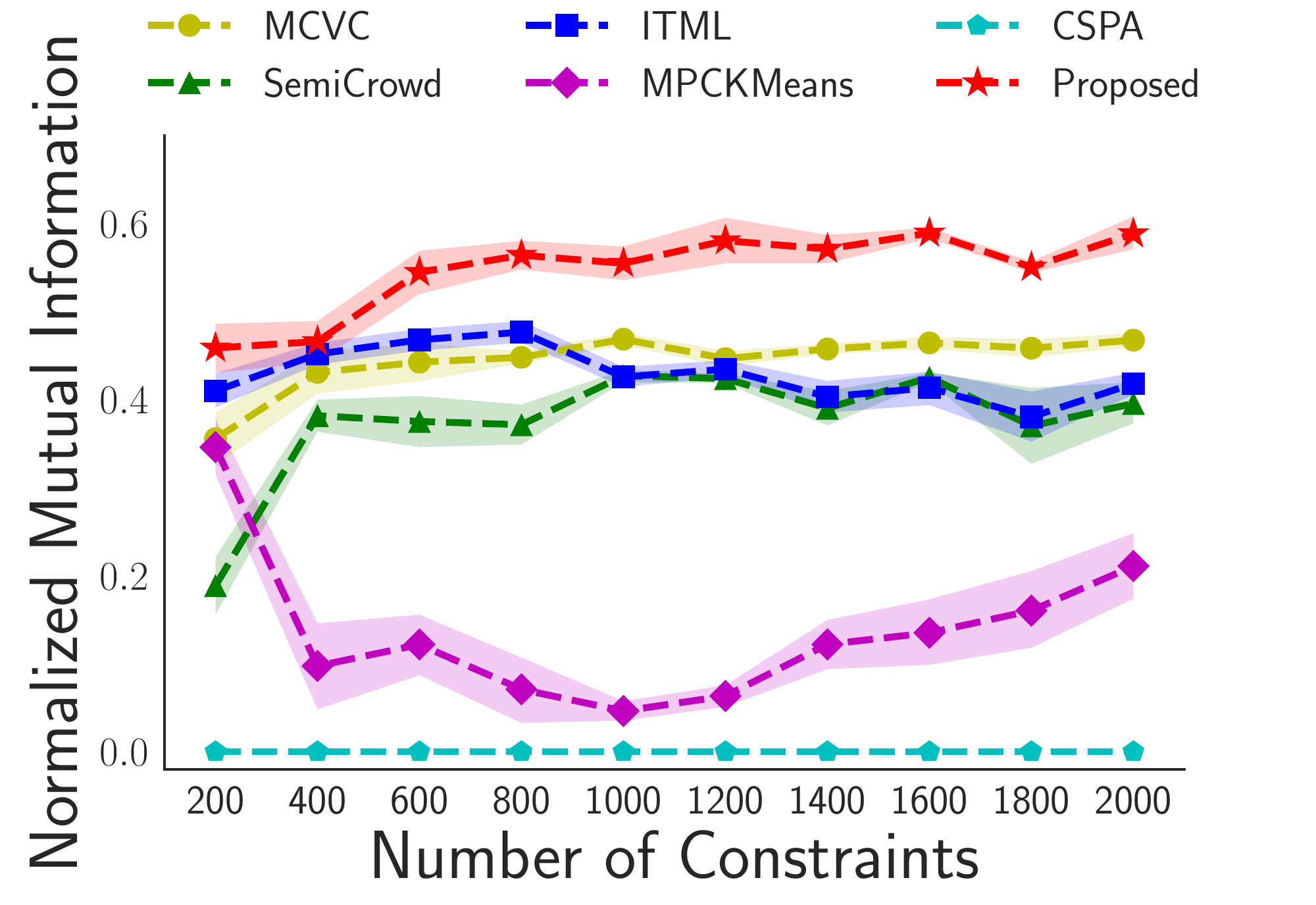}
            \caption{}
            \label{fig:face100}
        \end{subfigure}
        \hfill
        \begin{subfigure}[t]{.32\textwidth}
            \centering
            \includegraphics[height=3.3cm]{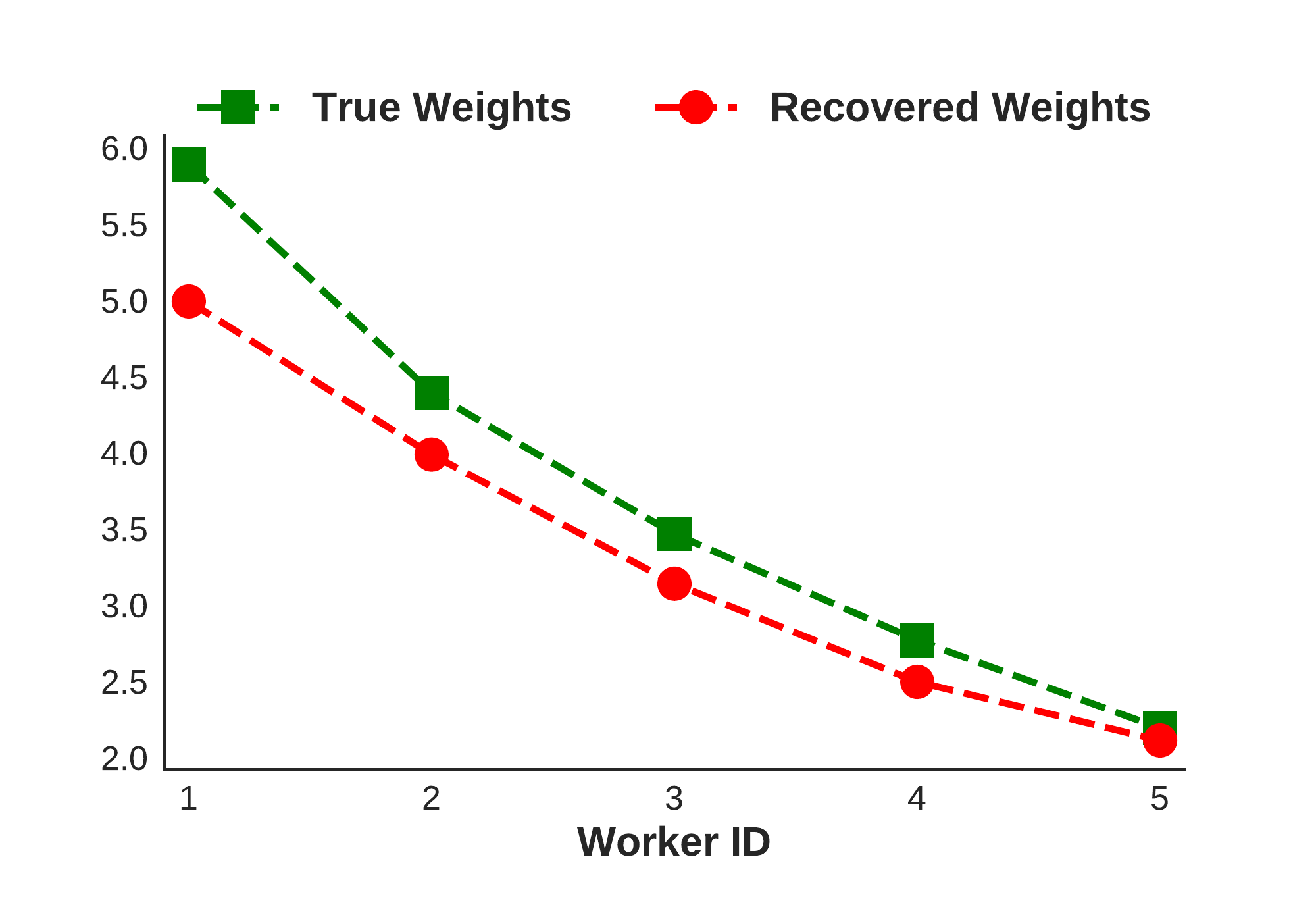}
            \caption{}
            \label{fig:pose100faceweight}
        \end{subfigure}
        \caption{Comparison to baselines: (a) Face: All the data points are annotated; (b) Face: Only 100 data points are annotated; (c) True accuracies are set to $\balpha=\bbeta=[0.95,0.9,0.85, 0.8,0.75]$. The green line is the true weights of each worker and the red line is the estimated weights by our model.} \vspace{-0.4cm}
        \label{fig:face}
    \end{figure}

    We test on Face dataset~\cite{Dua:2017}, containing 640 face images from 20 people with different poses (straight, left, right, up). The ground-truth clustering is based on the poses. The original image has 960 pixels. To speed up training, baseline methods apply Principle Component Analysis (PCA) and keep 20 components. For fair comparison, we test the proposed SCDC on the features after PCA. 
    Fig.~\ref{fig:face} plots the mean and standard deviation of NMI scores in 10 different runs for each fixed number of constraints. In Fig.~\ref{fig:face640}, the annotations are randomly generated on the whole dataset. We observe that our method consistently outperforms all competing methods, demonstrating that the clustering benefits from the joint generative modeling of inputs and annotations.
    
    \textbf{Annotations on a subset.}
    To illustrate the benefits of our method in the situation where only a small part of data points are annotated, we simulate noisy annotations on only 100 images. Fig.~\ref{fig:face100} shows the results of 100 annotated images. Our method exploits more structure information in the unlabeled data and shows notable improvements over all competing methods. 
    
    \textbf{Recover worker behaviors.}
    For each worker $m$, our model estimates the different accuracies $\alpha_m$ and $\beta_m$. We can derive from \cref{equ:ann} that the annotations of each worker $m$ are weighted by $\log \frac{\alpha_m}{1-\alpha_m} + \log \frac{\beta_m}{1-\beta_m}$, which means workers with higher accuracies are more reliable and will be weighted higher. We plot the weights of 5 workers in the Face experiments in Fig.~\ref{fig:pose100faceweight}.
    
    \vspace{-0.05cm}
    \subsection{End-to-end training with raw images}
\begin{figure}[t]
        \begin{center}
        \begin{subfigure}[b]{.67\textwidth}
            \centering
            \begin{tabular}{lc}
                \footnotesize Epoch 1 & \raisebox{-.4\height}{\Includegraphics[height=1.6cm]{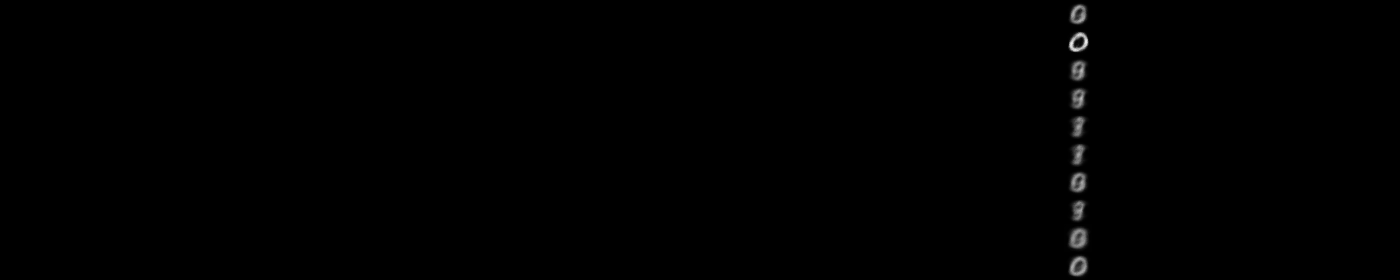}} \\
                \footnotesize Epoch 7 & \raisebox{-.4\height}{\Includegraphics[height=1.6cm]{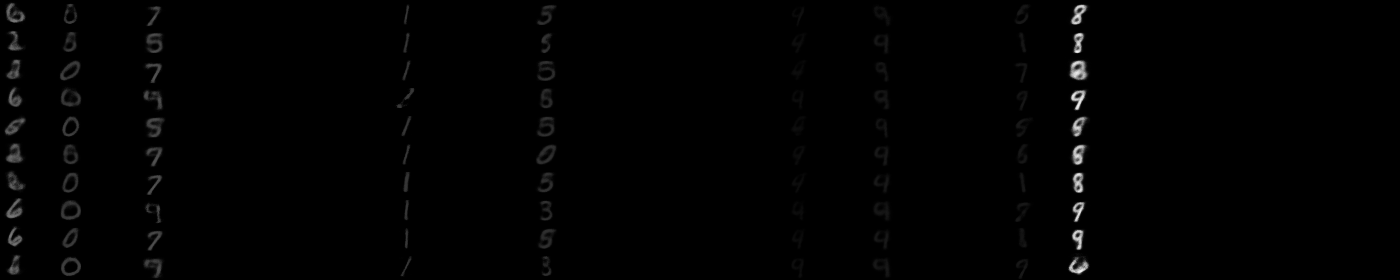}} \\
                \footnotesize Epoch 25 & \raisebox{-.4\height}{\Includegraphics[height=1.6cm]{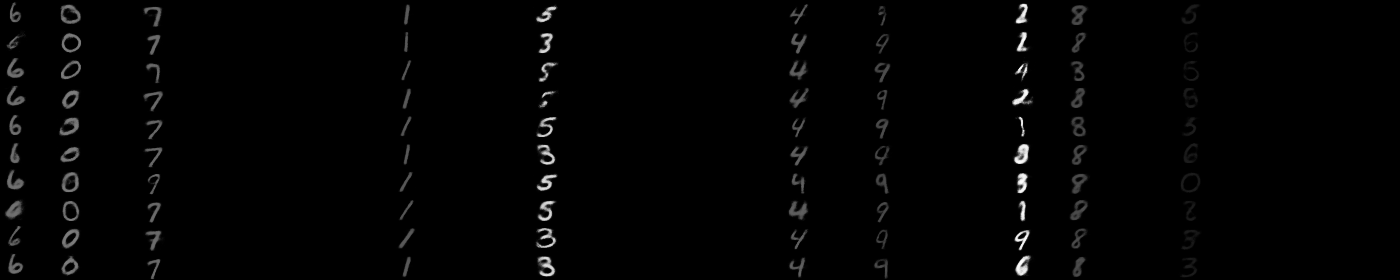}} \\
                \footnotesize Epoch 200 & \raisebox{-.4\height}{\Includegraphics[height=1.6cm]{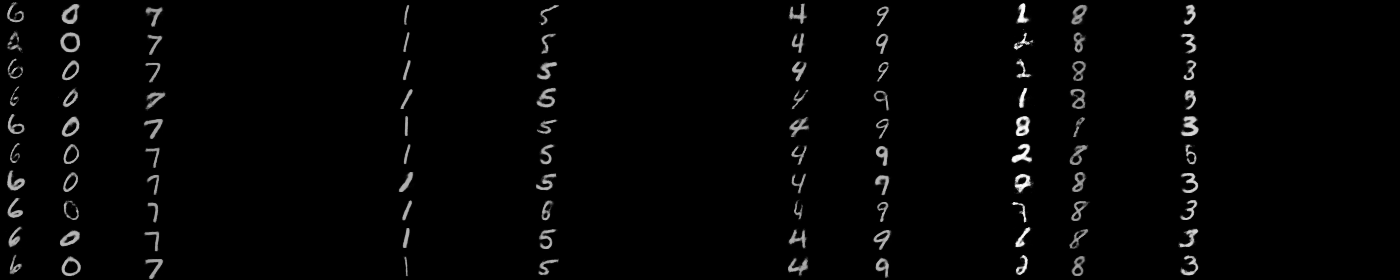}}
            \end{tabular}
            \caption{}
            \label{fig:mnist-auto}
        \end{subfigure}
        \hskip 0.46in
        \begin{subfigure}[b]{.23\textwidth}
            \centering
            \includegraphics[height=3cm]{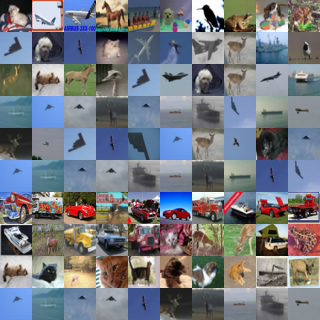}
            
            \vspace{3.4ex}
            
            \includegraphics[height=3cm]{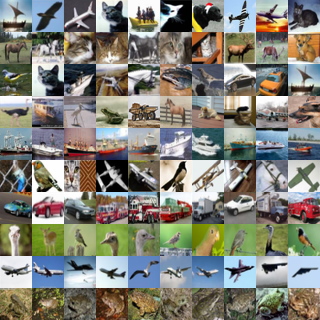}
            \caption{}
            \label{fig:cifar}
        \end{subfigure}
    \end{center}
        \caption{(a) MNIST: visualization of generated random samples of 50 clusters during training BayesSCDC. Each column represents a cluster, whose inferred proportion ($\pi_k$) is reflected by brightness; (b) Clustering results on CIFAR-10: (top) unsupervised; (bottom) with noisy annotations.} \vspace{-0.4cm}
    \end{figure}

    \paragraph{MNIST}
    As mentioned earlier, an important feature of DGMs is that they can directly model raw data, such as images. To verify this, we experiment with the MNIST dataset of digit images, which includes 60k training images from handwritten digits 0-9. 
    We collect crowdsourced annotations from $M=48$ workers and get 3276 %103-12=91 tasks
    annotations in total. 
     The two variants of our model (SCDC, BayesSCDC) are tested with or without annotations. For BayesSCDC, a non-informative prior $\mathrm{Beta}(1, 1)$ is placed over $\balpha, \bbeta$. For fair comparison, we also randomly sample the initial accuracy parameters $\balpha,\bbeta$ from $\mathrm{Beta}(1, 1)$ for SCDC. We average the results of 5 runs. In each run we randomly initialize the model for 10 times and pick the best result. All models are trained for 200 epochs with minibatch size of 128 for each random initialization.
    The results are shown in \Cref{tab:mnist}.  We can see that both models can effectively combine the information from the raw data and annotations, i.e., they worked reasonably well with only unlabeled data, and better when given noisy annotations on a subset of data. In terms of clustering accuracy and NMI, BayesSCDC outperforms SCDC. We believe that this is because the variational message passing algorithm used in BayesSCDC can effectively gather information from the crowdsourced annotations to form better variational approximations, as explained in \Cref{sec:natural}.
    Besides being more accurate, BayesSCDC is much faster because the computation cost caused by neural networks does not scales linearly with the number of clusters $K$ (50 in this case). In Fig.~\ref{fig:mnist-auto} we show that BayesSCDC is more flexible and automatically determines the number of mixture components during training.
    
    \begin{table}[t]
        \begin{center}
            \caption{Clustering performance on MNIST. The average time per epoch is reported.}
            \label{tab:mnist}
            \resizebox{\textwidth}{!} {
            \begin{tabular}{c|ccc|ccc}
                \toprule
                \multirow{2}{*}{\bf Method} &\multicolumn{3}{c|}{without annotations} & \multicolumn{3}{c}{with annotations} \\
                & \bf Accuracy & \bf NMI & {\bf Time} & \bf Accuracy & \bf NMI & {\bf Time} \\ 
                \midrule
                \makecell{SCDC}  & 65.92 $\pm$ 3.47 \% & 0.6953 $\pm$ 0.0167 &  177.3s & 81.87 $\pm$ 3.86\% & 0.7657 $\pm$ 0.0233 & 201.7s\\
                \midrule
                \makecell{BayesSCDC}  & 77.64 $\pm$ 3.97 \% & 0.7944 $\pm$ 0.0178 & 11.2s & \textbf{84.24 $\pm$ 5.52\%} & \textbf{0.8120 $\pm$ 0.0210} & 16.4s \\
                \bottomrule
            \end{tabular}}
        \end{center}\vspace{-0.2cm}
    \end{table}

    % For SCDC, we set the initial accuracy parameters of workers to 0.5-0.9. For BayesSCDC, a non-informative prior $\mathrm{Beta}(1, 1)$ is placed over them. The results are shown in Fig.~\ref{fig:mnist} (left). We can see that both models can effectively combine the information from the unsupervised data and annotations, and BayesSCDC clearly outperforms SCDC in terms of clustering accuracy. As mentioned in \Cref{sec:bayes}, BayesSCDC is much faster because the computation cost caused by neural networks does not scales linearly with the number of clusters $K$ (50 in this case).
    % In Fig.~\ref{fig:mnistwithoutanns} and~\ref{fig:mnistwithanns} we plot random generated images from the learned model. Each row is from the same cluster. We can see that without any annotations, the generative model mistakes similar digits such as 4 and 9, 3 and 5.
    
    \paragraph{CIFAR-10}
    We also conduct experiments with real crowdsourced labels on more complex natural images, i.e., {CIFAR-10}. Using the same crowdsourcing scheme, we collect 8640 % 291-51 unique assignments
    noisy annotations from 32 web workers on a subset of randomly sampled 4000 images. We apply SCDC with/without annotations for 5 runs of random initializations. 
%    $p(\bo|\bx)$ is parameterized by a deep stack of residual blocks.
    SCDC without annotations failed with NMI score 0.0424 $\pm$ 0.0119 and accuracy 14.23 $\pm$ 0.69\% among 5 runs. But the NMI score achieved by SCDC with noisy annotations is 0.5549 $\pm$ 0.0028 and the accuracy is 50.09 $\pm$ 0.08\%. The clustering results on test dataset are shown in Fig.~\ref{fig:cifar}. We plot 10 test samples with the largest probability for each cluster. More experiment details and discussions could be found in the supplementary material.
    \vspace{-0.2cm}
    \section{Conclusion}
    \vspace{-0.1cm}
    In this paper, we proposed a semi-crowdsourced clustering model based on deep generative models and its fully Bayesian version. We developed fast (natural-gradient) stochastic variational inference algorithms for them. The resulting method can jointly model the crowdsourced labels, worker behaviors, and the (un)annotated items. Experiments have demonstrated that the proposed method outperforms previous competing methods on standard benchmark datasets. Our work also provides general guidelines on how to incorporate DGMs to statistical relational models, where the proposed inference algorithm can be applied under a broader context.
  
    \section*{Acknowledgement}
    Yucen Luo would like to thank Matthew Johnson for helpful discussions on the SVAE algorithm~\citep{johnson2016composing}, and Yale Chang for sharing the code of the UCI benchmark experiments. We thank the anonymous reviewers for feedbacks that greatly improved the paper. This work was supported by the National Key Research and Development Program of China (No. 2017YFA0700904), NSFC Projects (Nos. 61620106010, 61621136008, 61332007), Beijing NSF Project (No. L172037), Tiangong Institute for Intelligent Computing, NVIDIA NVAIL Program, and the projects from Siemens, NEC and Intel.
    
    {
        \small
        \bibliographystyle{plain}
        \bibliography{ref}

\begin{thebibliography}{10}

\bibitem{bilenko2004integrating}
Mikhail Bilenko, Sugato Basu, and Raymond~J Mooney.
\newblock Integrating constraints and metric learning in semi-supervised
  clustering.
\newblock In {\em Proceedings of the twenty-first international conference on
  Machine learning}, page~11. ACM, 2004.

\bibitem{chandola2009anomaly}
Varun Chandola, Arindam Banerjee, and Vipin Kumar.
\newblock Anomaly detection: A survey.
\newblock {\em ACM computing surveys (CSUR)}, 41(3):15, 2009.

\bibitem{chang2017multiple}
Yale Chang, Junxiang Chen, Michael~H Cho, Peter~J Castaldi, Edwin~K Silverman,
  and Jennifer~G Dy.
\newblock Multiple clustering views from multiple uncertain experts.
\newblock In {\em International Conference on Machine Learning}, pages
  674--683, 2017.

\bibitem{davis2007information}
Jason~V Davis, Brian Kulis, Prateek Jain, Suvrit Sra, and Inderjit~S Dhillon.
\newblock Information-theoretic metric learning.
\newblock In {\em Proceedings of the 24th international conference on Machine
  learning}, pages 209--216. ACM, 2007.

\bibitem{dawid1979maximum}
Alexander~Philip Dawid and Allan~M Skene.
\newblock Maximum likelihood estimation of observer error-rates using the em
  algorithm.
\newblock {\em Applied Statistics}, pages 20--28, 1979.

\bibitem{deng2009imagenet}
Jia Deng, Wei Dong, Richard Socher, Li-Jia Li, Kai Li, and Li~Fei-Fei.
\newblock Imagenet: A large-scale hierarchical image database.
\newblock In {\em Computer Vision and Pattern Recognition, 2009. CVPR 2009.
  IEEE Conference on}, pages 248--255. IEEE, 2009.

\bibitem{Dua:2017}
Dua Dheeru and Efi Karra~Taniskidou.
\newblock {UCI} machine learning repository, 2017.

\bibitem{gomes2011crowdclustering}
Ryan~G Gomes, Peter Welinder, Andreas Krause, and Pietro Perona.
\newblock Crowdclustering.
\newblock In {\em Advances in neural information processing systems}, pages
  558--566, 2011.

\bibitem{gopalan2012scalable}
Prem~K Gopalan, Sean Gerrish, Michael Freedman, David~M Blei, and David~M
  Mimno.
\newblock Scalable inference of overlapping communities.
\newblock In {\em Advances in Neural Information Processing Systems}, pages
  2249--2257, 2012.

\bibitem{he2016deep}
Kaiming He, Xiangyu Zhang, Shaoqing Ren, and Jian Sun.
\newblock Deep residual learning for image recognition.
\newblock In {\em Proceedings of the IEEE conference on computer vision and
  pattern recognition}, pages 770--778, 2016.

\bibitem{hoffman2013stochastic}
Matthew~D Hoffman, David~M Blei, Chong Wang, and John Paisley.
\newblock Stochastic variational inference.
\newblock {\em The Journal of Machine Learning Research}, 14(1):1303--1347,
  2013.

\bibitem{howe2006rise}
Jeff Howe.
\newblock The rise of crowdsourcing.
\newblock {\em Wired magazine}, 14(6):1--4, 2006.

\bibitem{johnson2016composing}
Matthew Johnson, David~K Duvenaud, Alex Wiltschko, Ryan~P Adams, and Sandeep~R
  Datta.
\newblock Composing graphical models with neural networks for structured
  representations and fast inference.
\newblock In {\em Advances in neural information processing systems}, pages
  2946--2954, 2016.

\bibitem{kingma2014semi}
Diederik~P Kingma, Shakir Mohamed, Danilo~Jimenez Rezende, and Max Welling.
\newblock Semi-supervised learning with deep generative models.
\newblock In {\em Advances in Neural Information Processing Systems}, pages
  3581--3589, 2014.

\bibitem{kingma2013auto}
Diederik~P Kingma and Max Welling.
\newblock Auto-encoding variational bayes.
\newblock {\em arXiv preprint arXiv:1312.6114}, 2013.

\bibitem{lin2018variational}
Wu~Lin, Mohammad~Emtiyaz Khan, and Nicolas Hubacher.
\newblock Variational message passing with structured inference networks.
\newblock In {\em International Conference on Learning Representations}, 2018.

\bibitem{Luo_2018_CVPR}
Yucen Luo, Jun Zhu, Mengxi Li, Yong Ren, and Bo~Zhang.
\newblock Smooth neighbors on teacher graphs for semi-supervised learning.
\newblock In {\em The IEEE Conference on Computer Vision and Pattern
  Recognition}, 2018.

\bibitem{raykar2010learning}
V.~C. Raykar, S.~Yu, L.~H. Zhao, G.~H. Valadez, C.~Florin, L.~Bogoni, and
  L.~Moy.
\newblock Learning from crowds.
\newblock {\em JMLR}, 11:1297--1322, 2010.

\bibitem{rezende2016one}
Danilo~J Rezende, Shakir Mohamed, Ivo Danihelka, Karol Gregor, and Daan
  Wierstra.
\newblock One-shot generalization in deep generative models.
\newblock In {\em Proceedings of the 33rd International Conference on
  International Conference on Machine Learning-Volume 48}, pages 1521--1529.
  JMLR. org, 2016.

\bibitem{shi2000normalized}
Jianbo Shi and Jitendra Malik.
\newblock Normalized cuts and image segmentation.
\newblock {\em IEEE Transactions on pattern analysis and machine intelligence},
  22(8):888--905, 2000.

\bibitem{zhusuan2017}
Jiaxin Shi, Jianfei. Chen, Jun Zhu, Shengyang Sun, Yucen Luo, Yihong Gu, and
  Yuhao Zhou.
\newblock Zhu{S}uan: A library for {B}ayesian deep learning.
\newblock {\em arXiv preprint arXiv:1709.05870}, 2017.

\bibitem{strehl2002cluster}
Alexander Strehl and Joydeep Ghosh.
\newblock Cluster ensembles---a knowledge reuse framework for combining
  multiple partitions.
\newblock {\em Journal of machine learning research}, 3(Dec):583--617, 2002.

\bibitem{Tian2015Max}
Tian Tian and Jun Zhu.
\newblock Max-margin majority voting for learning from crowds.
\newblock In {\em Advances in Neural Information Processing Systems}, pages
  1621--1629, 2015.

\bibitem{Vinayak2016Crowdsourced}
Ramya~Korlakai Vinayak and Babak Hassibi.
\newblock Crowdsourced clustering: Querying edges vs triangles.
\newblock In {\em Neural Information Processing System}, 2016.

\bibitem{welinder2010multidimensional}
Peter Welinder, Steve Branson, Pietro Perona, and Serge~J Belongie.
\newblock The multidimensional wisdom of crowds.
\newblock In {\em Advances in neural information processing systems}, pages
  2424--2432, 2010.

\bibitem{winn2005variational}
John Winn and Christopher~M Bishop.
\newblock Variational message passing.
\newblock {\em Journal of Machine Learning Research}, 6(Apr):661--694, 2005.

\bibitem{wiwie2015comparing}
Christian Wiwie, Jan Baumbach, and Richard R{\"o}ttger.
\newblock Comparing the performance of biomedical clustering methods.
\newblock {\em Nature methods}, 12(11):1033--1038, 2015.

\bibitem{xie2016unsupervised}
Junyuan Xie, Ross Girshick, and Ali Farhadi.
\newblock Unsupervised deep embedding for clustering analysis.
\newblock In {\em International conference on machine learning}, pages
  478--487, 2016.

\bibitem{xing2003distance}
Eric~P Xing, Michael~I Jordan, Stuart~J Russell, and Andrew~Y Ng.
\newblock Distance metric learning with application to clustering with
  side-information.
\newblock In {\em Advances in neural information processing systems}, pages
  521--528, 2003.

\bibitem{yi2012semi}
Jinfeng Yi, Rong Jin, Shaili Jain, Tianbao Yang, and Anil~K Jain.
\newblock Semi-crowdsourced clustering: Generalizing crowd labeling by robust
  distance metric learning.
\newblock In {\em Advances in neural information processing systems}, pages
  1772--1780, 2012.

\bibitem{zhouaggregating}
Dengyong Zhou, Qiang Liu, John Platt, and Christopher Meek.
\newblock Aggregating ordinal labels from crowds by minimax conditional
  entropy.
\newblock In {\em Proceedings of the 31th International Conference on Machine
  Learning, {ICML} 2014, Beijing, China, 21-26 June 2014}, pages 262--270,
  2014.

\end{thebibliography}
    }
    
    \clearpage
    \appendix
    \section{Derivations}
    \label{app:deriv}
    \subsection{Natural parameters and sufficient statistics}
    $$
    \bmeta_{\bpi} =\begin{bmatrix}
    \alpha_1 - 1 \\
    \vdots \\
    \alpha_K - 1
    \end{bmatrix},
    \bmeta_{\bz}(\bpi) =\begin{bmatrix}
    \log \pi_1 \\
    \vdots \\
    \log \pi_K \end{bmatrix}, 
    (\bmeta_{\bmu,\bSigma})^{(k)}= 
    \begin{bmatrix}
    \kappa\bbm \\
    \mathrm{vec}(\bS + \kappa\bbm\bbm^\top) \\
    \kappa \\
    \nu + d + 2
    \end{bmatrix},
    \bmeta_\bx(\bmu, \bSigma) =\begin{bmatrix} 
    \bSigma^{-1}\bmu \\ 
    \mathrm{vec}(-\frac{1}{2}\bSigma^{-1})
    \end{bmatrix}.
    $$
    
    $$
    \bt(\bpi) =\begin{bmatrix}
    \log \pi_1\\
    \vdots\\
    \log \pi_K
    \end{bmatrix},
    \bt(\bz)=\begin{bmatrix}
    z_1\\
    \vdots\\
    z_K\end{bmatrix}, 
    \bt(\bmu,\bSigma)^{(k)} = \begin{bmatrix}
    \bSigma_k^{-1}\bmu_k\\
    \mathrm{vec}{(-\frac{1}{2}\bSigma_k^{-1})}\\
    -\frac{1}{2}\bmu_k^\top\bSigma_k^{-1}\bmu_k \\
    -\frac{1}{2}\ln |\bSigma_k|
    \end{bmatrix},
    \bt(\bx) =\begin{bmatrix} 
    \bx \\ 
    \mathrm{vec}( \bx\bx^\top )\end{bmatrix}.
    $$
    
    $$
    (\bmeta_{\balpha})^{(m)} = \begin{bmatrix}
    \tau_{\alpha_m^1}-1 \\
    \tau_{\alpha_m^2}-1
    \end{bmatrix}, \quad
    \bt(\balpha)^{(m)} = \begin{bmatrix}
    \log\alpha_m \\
    \log (1 - \alpha_m)
    \end{bmatrix},\quad\text{$\bmeta_{\bbeta}$ and $\bt(\bbeta)$ are similar.}
    $$
    
    Note that $\bt(\bpi) = \bmeta_{\bz}(\bpi)$, $\bmeta_{\bx}(\bmu, \bSigma) = \bt(\bmu, \bSigma)[:2]$, where the conjugacy comes.
    In practice, we parameterized the unnormalized version of $
    \tilde{\bmeta_{\bz}}(\bpi) = \begin{bmatrix}
    \log \pi_1 + c \\
    \vdots \\
    \log \pi_K + c \end{bmatrix},
    $ since it is unconstrained. The update rule for $\tilde{\bmeta_{\bz}}(\bpi)$ is the same with $\bmeta_{\bz}(\bpi)$ due to their constant difference.
    \subsection{Expected sufficient statistics}
    $$
    \mathbb{E}_{q(\bpi)} \bt(\bpi) = \begin{bmatrix}
    \psi(\alpha_1)\\
    \vdots \\
    \psi(\alpha_K)
    \end{bmatrix} - \psi\left(\sum_{k=1}^K\alpha_k\right), \quad
    \mathbb{E}_{q(\bmu, \bSigma)}\bt(\bmu, \bSigma)^{(k)} = \begin{bmatrix}
    \nu \bS^{-1}\bbm \\
    \mathrm{vec}(-\frac{1}{2}\nu\bS^{-1}) \\
    -\frac{1}{2}(\kappa^{-1}d + \nu\bbm^\top\bS^{-1}\bbm) \\
    \frac{1}{2}\left(\psi_d(\frac{\nu}{2}) + d\ln 2 - \ln |\bS|\right)
    \end{bmatrix},
    $$
    where $\psi_d(\frac{\nu}{2}) = \sum_{i=1}^d\psi(\frac{\nu + 1- i}{2})$, $\psi$ is the digamma function.
    $$
    \mathbb{E}_{q(\bz)}\bt(\bz) = \begin{bmatrix}
    \pi_1 \\
    \vdots \\
    \pi_K
    \end{bmatrix}, \quad
    \mathbb{E}_{q(\bx)}\bt(\bx) = \begin{bmatrix}
    \bmu, \\
    \mathrm{vec}(\bSigma + \bmu\bmu^\top)
    \end{bmatrix}
    $$
    $$
    \mathbb{E}_{q(\balpha)}\bt(\balpha)^{(m)} = \begin{bmatrix}
    \psi(\tau_{\alpha_m^1}) \\
    \psi(\tau_{\alpha_m^2})
    \end{bmatrix} - \psi(\tau_{\alpha_m^1} + \tau_{\alpha_m^2}),\quad \text{$\mathbb{E}_{q(\bbeta)}\bt(\bbeta)$ is similar.}
    $$
    
    \subsection{Log partition function}
    $$
    \ln Z(\bmeta_{\bpi}) = \sum_{k=1}^K\ln\Gamma(\alpha_k) - \ln\Gamma\left(\sum_{k=1}^K\alpha_k\right), \quad
    \ln Z(\bmeta_{\bmu, \bSigma}) = \frac{\nu}{2}(d\ln 2 - \ln |S|) + \ln \Gamma_d\left(\frac{\nu}{2}\right) - \frac{d}{2}\ln \kappa,
    $$
    where $\Gamma_d(\frac{\nu}{2}) = \pi^{d(d - 1)/4}\prod_{i=1}^d\Gamma\left(\frac{\nu + 1 - i}{2}\right)$, $\Gamma$ is the Gamma function.
    $$
    \ln Z(\bmeta_{\bz}(\bpi)) = 0, \quad
    \ln Z(\bmeta_{\bx}(\bmu, \bSigma)) = \frac{1}{2}\left(\bmu^\top\bSigma^{-1}\bmu + \ln |\bSigma|\right) = -\mathbf{1}^\top\{\bt(\bmu, \bSigma)[2:]\}.
    $$
    $$
    \ln Z(\bmeta_{\balpha}) = \ln\Gamma(\tau_{\alpha^1}) + \ln\Gamma(\tau_{\alpha^2}) - \ln\Gamma(\tau_{\alpha^1} + \tau_{\alpha^2}).
    $$
    \subsection{Variational message passing for local parameters}
    The local update for $\bx_i$:
    \begin{equation}
    \bmeta_{\bx_i}^*=\mathbb{E}_{q(\bmu,\bSigma)}[\bt(\bmu,\bSigma)[:2]]^\top\mathbb{E}_{q(\bz_i)}\bt(\bz_i)\!+\!r(\bo_i;\bphi).
    \end{equation}
    The local update for $\bz_i$:
    \begin{equation*}
    \bmeta_{\bz_i}^* =  \mathbb{E}_{q(\bpi)}\bt(\bpi) + \mathbb{E}_{q(\bmu,\bSigma)}\left[\bt(\bmu,\bSigma)\right]^\top\mathbb{E}_{q(\bx_i)}\left[(\bt(\bx_i), \mathbf{1})\right] + \sum_{m\!=\!1}^M\sum_{j=1}^N w_{ij}^{(m)}\mathbb{E}_{q(\bz_j)}[\bt(\bz_j)].
    \end{equation*}
    where
    $w_{ij}^{(m)} = I_{ij}^{(m)}\mathbb{E}_{q(\balpha, \bbeta)}\left[\ln\frac{1-\alpha_m}{\beta_m}+L_{ij}^{(m)}\left(\ln\frac{\alpha_m}{1-\alpha_m} + \ln\frac{\beta_m}{1-\beta_m}\right)\right]$.
    \subsection{The final objective}
    The final objective:
    \begin{align}
    \mathcal{J} (\bmeta_{\bTheta};\bphi,\bgamma)\triangleq &~\mathcal{L}(\bmeta_{\bTheta},\bmeta_{\bZ}^*(\bmeta_{\bTheta},\bphi), \bmeta_{\bX}^*(\bmeta_{\bTheta},\bphi),\bgamma) \notag\\
    =&~\mathbb{E}_{\opt q(\bX)}\! \log p(\bO|\bX;\bgamma) + \mathbb{E}_{q(\balpha,\bbeta)\opt q(\bZ)} \log p(\bL^{(1:M)}|\bZ,\balpha,\bbeta) \notag\\
    & - \mathbb{E}_{q(\bpi,\bmu,\bSigma)}\KL (\opt q(\bZ)\opt q(\bX)\|p(\bZ|\bpi)p(\bX|\bmu,\bSigma,\bZ)) \notag\\
    & - \KL(q(\bpi,\bmu,\bSigma, \balpha, \bbeta) \| p(\bpi,\bmu,\bSigma, \balpha, \bbeta)).
    \end{align}
    The annotation likelihood term:
    \begin{align}
    &\mathbb{E}_{q(\balpha, \bbeta)q^*(\bZ)}\log p(\bL^{(1:M)}|\bZ,\balpha,\bbeta) = \notag\\
    &\quad\sum_{m=1}^M\sum_{1\leq i<j\leq N}\left[w_{ij}^{(m)}\mathbb{E}_{q(\bz_i)}[\bt(\bz_i)]^\top\mathbb{E}_{q(\bz_j)}\bt(\bz_j) + I_{ij}^{(m)}\mathbb{E}_{q(\bbeta)}\left(L_{ij}^{(m)}\ln\frac{1 - \beta_m}{\beta_m} + \ln\beta_m\right)\right].
    \end{align}
    The local KL divergence term:
    \begin{align}
    &~\mathbb{E}_{q(\bpi,\bmu,\bSigma)}\KL (\opt q(\bZ)\opt q(\bX)\|p(\bZ|\bpi)p(\bX|\bmu,\bSigma,\bZ)) \notag\\
    &=~\mathbb{E}_{q(\bpi)}\mathrm{KL}(q^*(\bZ)\|p(\bZ|\bpi)) + \mathbb{E}_{q(\bmu,\bSigma)q^*(\bZ)}\mathrm{KL}(q^*(\bX)\|p(\bX|\bmu,\bSigma,\bZ)) \notag\\
    &=~\sum_{i=1}^N\left\{ \mathbb{E}_{q(\bpi)}\mathrm{KL}(q^*(\bz_i)\|p(\bz_i|\bpi)) + \mathbb{E}_{q(\bmu,\bSigma)q^*(\bz_i)}\mathrm{KL}(q^*(\bx_i)\|p(\bx_i|\bmu,\bSigma,\bz_i))\right\}.
    \end{align}
    \begin{align*}
    \mathrm{KL}(q^*(\bz_i)\|p(\bz_i|\bpi))&=\langle\bmeta_{\bz_i}^* - \bmeta_{\bz_i}(\bpi), \mathbb{E}_{q^*(\bz_i)}\bt(\bz_i)\rangle \\
    \mathbb{E}_{q(\bpi)}\mathrm{KL}(q^*(\bz_i)\|p(\bz_i|\bpi)) &= \langle\bmeta_{\bz_i}^* - \mathbb{E}_{q(\bpi)}\bt(\bpi), \mathbb{E}_{q^*(\bz_i)}\bt(\bz_i)\rangle.
    \end{align*}
    \begin{equation}
    \mathrm{KL}(q^*(\bx_i)\|p(\bx_i|\bmu_k,\bSigma_k)) = \langle\bmeta_{\bx_i}^* - \bmeta_{\bx_i}(\bmu,\bSigma)^{(k)}, \mathbb{E}_{q^*(\bx_i)}\bt(\bx_i)\rangle - \left[\ln Z(\bmeta_{\bx_i}^*) - \ln Z(\bmeta_{\bx_i}(\bmu,\bSigma)^{(k)})\right]
    \end{equation}
    \begin{align*}
    \mathbb{E}_{q(\bmu,\bSigma)q^*(\bz_i)}\mathrm{KL}(q^*(\bx_i)\|p(\bx_i|\bmu,\bSigma,\bz_i))
    %        &\quad=\mathbb{E}_{q(\bmu,\bSigma)}\sum_{k=1}^K\bpi_k\left\{\langle\bmeta_{\bx_i}^* - \bmeta_{\bx_i}(\bmu_k,\bSigma_k), \mathbb{E}_{q^*(\bx_i)}\bt(\bx_i)\rangle - \left[\ln Z(\bmeta_{\bx_i}^*) - \ln Z(\bmeta_{\bx_i}(\bmu_k,\bSigma_k))\right]\right\} \\
    %    &\quad=\sum_{k=1}^K\bpi_k\left\{\langle\bmeta_{\bx_i}^* - \mathbb{E}_{q(\bmu,\bSigma)}[\bmeta_{\bx_i}(\bmu_k,\bSigma_k)], \mathbb{E}_{q^*(\bx_i)}\bt(\bx_i)\rangle - \left[\ln Z(\bmeta_{\bx_i}^*) - \mathbb{E}_{q(\bmu,\bSigma)}\ln Z(\bmeta_{\bx_i}(\bmu_k,\bSigma_k))\right]\right\} \\
    =&\langle\bmeta_{\bx_i}^* - \mathbb{E}_{q(\bmu,\bSigma)}[\bmeta_{\bx_i}(\bmu,\bSigma)]^\top\mathbb{E}_{q^*(\bz_i)}\bt(\bz_i), \mathbb{E}_{q^*(\bx_i)}\bt(\bx_i)\rangle \\
    &-\left\{\ln Z(\bmeta_{\bx_i}^*) - \mathbb{E}_{q(\bmu,\bSigma)}[\ln Z(\bmeta_{\bx_i}(\bmu,\bSigma))]^\top\mathbb{E}_{q*(\bz_i)}\bt(\bz_i) \right\} \\
    %    &=\langle\bmeta_{\bx_i}^*, \mathbb{E}_{q^*(\bx_i)}\bt(\bx_i)\rangle - \ln Z(\bmeta_{\bx_i}^*) - \langle\mathbb{E}_{q(\bmu,\bSigma)}[\bt(\bmu,\bSigma)]^\top\mathbb{E}_{q^*(\bz_i)}\bt(\bz_i), (\mathbb{E}_{q^*(\bx_i)}\bt(\bx_i), \mathbf{1})\rangle \\
    =&\langle\bmeta_{\bx_i}^* - \mathbb{E}_{q(\bmu,\bSigma)}[\bt(\bmu,\bSigma)[:2]]^\top\mathbb{E}_{q^*(\bz_i)}\bt(\bz_i), \mathbb{E}_{q^*(\bx_i)}\bt(\bx_i)\rangle \\
    &- \left\{\ln Z(\bmeta_{\bx_i}^*) + \mathbf{1}^\top\mathbb{E}_{q(\bmu,\bSigma)}[\bt(\bmu,\bSigma)[2:]]^\top\mathbb{E}_{q*(\bz_i)}\bt(\bz_i) \right\}.
    \end{align*}
    The global KL divergence term:
    \begin{align}
    \mathrm{KL}(q(\bpi)\|p(\bpi)) = \langle\bmeta_{\bpi} - \bmeta^0_{\bpi}, \mathbb{E}_{q(\bpi)}\bt(\bpi)\rangle - \left[\ln Z(\bmeta_{\bpi}) - \ln Z(\bmeta_{\bpi}^0)\right].
    \end{align}
    \begin{align}
    \mathrm{KL}(q(\bmu, \bSigma)\|p(\bmu, \bSigma)) &= \sum_{k=1}^K\left\{{\langle\bmeta_{\bmu,\bSigma}}^{(k)} - {\bmeta^0_{\bmu,\bSigma}}^{(k)}, \mathbb{E}_{q(\bmu,\bSigma)}\bt(\bmu, \bSigma)^{(k)}\rangle\right. \notag \\
    &\quad- \left.\left[\ln Z({\bmeta_{\bmu,\bSigma}}^{(k)}) - \ln Z({\bmeta_{\bmu,\bSigma}^0}^{(k)})\right]\right\}.
    \end{align}
    \begin{align}
    \mathrm{KL}(q(\balpha)\|p(\balpha)) = \langle\bmeta_{\balpha} - \bmeta^0_{\balpha}, \mathbb{E}_{q(\balpha)}\bt(\balpha)\rangle - \left[\ln Z(\bmeta_{\balpha}) - \ln Z(\bmeta_{\balpha}^0)\right], \quad\text{$\bbeta$ is similar}.
    \end{align}
    Global variational parameter updates:
    \begin{align}
    \tilde{\nabla}_{\bmeta_{\bpi}}\mathcal{J} &\approx \bmeta_{\bpi}^0 + \sum_{i=1}^N\mathbb{E}_{q(\bz_i)}\bt(\bz_i) - \bmeta_{\bpi}, \\
    \tilde{\nabla}_{\bmeta_{\bmu, \bSigma}}\mathcal{J} &\approx \bmeta_{\bmu,\bSigma}^0 + \sum_{i=1}^N \left[\mathbb{E}_{q(\bx_i)}(\bt(\bx_i), \mathbf{1})\right]^\top \mathbb{E}_{q(\bz_i)}\bt(\bz_i) - \bmeta_{\bmu,\bSigma}, \\
    \tilde{\nabla}_{\bmeta_{\balpha}^{(m)}}\mathcal{J} &= (\bmeta_{\balpha}^0)^{(m)} + \frac{1}{2}\sum_{i,j=1}^N I_{ij}^{(m)}\mathbb{E}_{q(\bz_i)}[\bt(\bz_i)]^\top\mathbb{E}_{q(\bz_j)}\bt(\bz_j)\begin{bmatrix}
    L^{(m)}_{ij} \\
    1 - L^{(m)}_{ij}
    \end{bmatrix} -  \bmeta_{\balpha}^{(m)}, \\
    \tilde{\nabla}_{\bmeta_{\bbeta}^{(m)}}\mathcal{J} &= (\bmeta_{\bbeta}^0)^{(m)} + \frac{1}{2}\sum_{i,j=1}^N I_{ij}^{(m)}\left(1 - \mathbb{E}_{q(\bz_i)}[\bt(\bz_i)]^\top\mathbb{E}_{q(\bz_j)}\bt(\bz_j)\right)\begin{bmatrix}
    1 - L^{(m)}_{ij} \\
    L^{(m)}_{ij}
    \end{bmatrix} -  \bmeta_{\bbeta}^{(m)}.
    \end{align}
    \section{Experiment settings}
	\paragraph{Toy Pinwheel dataset} We follow the synthetic data generation process and hyperparameter settings in~\cite{johnson2016composing}. The latent dim of $\bx$ is set to $d = 2$ and initial number of clusters $K=15$.
	The Dirichlet prior for mixing coefficients $\bpi$: $\alpha_0 = 0.05/K$. The NIW prior $p(\bmu,\bSigma)$: concentration $\kappa=0.5$, location parameter $\bm{m} = \mathbf{0}$, scale matrix $\mathbf{S}=(d+\kappa)\mathbf{I}_d$ ($\mathbf{I}_d \in \mathbb{R}^{d\times d}$ is the identity matrix), degrees of freedom $\nu=d+\kappa$. The Beta prior hyperparameter $\tau_{\alpha_0^1}, \tau_{\alpha_0^2}, \tau_{\beta_0^1}, \tau_{\beta_0^2}$ for accuracy $(\balpha,\bbeta)$ are all set to 1.
	The hyperparameters in variational distributions of global variables are $\alpha_0 \sim \mathcal{U}(1,2), \kappa = 1, \bm{m} \sim \mathcal{N}(0,3), \mathbf{S}=(d+\kappa)\mathbf{I}_d,\nu=d+\kappa, \tau_{\alpha_0^1} = \tau_{\beta_0^1} =10, \tau_{\alpha_0^2} = \tau_{\beta_0^2} = 1$.
	 The networks of $p(\bo|\bx;\bgamma)$ and recognition networks $r(\bo;\bphi)$ are both MLPs with two hidden layers of 40 ReLU units. The networks are trained for 20 epochs with minibatch size $|B|$ of 50, momentum 0.9 and learning rate $1\times10^{-3}$. The learning rate for global variational parameters $\bmeta_\bTheta$ is $1\times 10^{-4}$. The batch size $|S|$ of annotations is computed by $|S| = N_a \times |B|/N$ to ensure that the annotations and the images are feed to the network proportionally.
    
    \paragraph{UCI dataset} The implementation of competing methods are based on the source code provided by MCVC~\cite{chang2017multiple}. In our proposed SCDC, the preprocessing of images and generation of annotations are the same as that in MCVC. The latent dim is set to $d=8$ and $K=4$. As for the network architecture, $p(\bo|\bx;\bgamma)$ and recognition networks $q(\bz_n|\bo_n;\bphi)q(\bx_n|\bz_n,\bo_n;\bphi)$ are simple MLPs with two hidden layers of 40 ReLU units. The networks are trained for 20 epochs using Adam optimizer with learning rate $1\times 10^{-3}$ and batch size $|B|=10$ . 
    
    \paragraph{MNIST} For both SCDC and BayesSCDC, $p(\bo|\bx)$ and recognition networks are MLPs with two hidden layers of 500 ReLU units. The latent dim of $\bx$ is $d=8$ and initial number of clusters $K =50$. Both two models are trained for 200 epochs. For SCDC, we use Adam optimizer with learning rate $1\times10^{-3}$ and minibatch size 128. For BayesSCDC, the hyperparameter settings in the priors and variational distributions are all the same as that in the Pinwheel dataset.
    % except that the Dirichlet prior for mixing coefficients $p(\bpi)$ has $\alpha_0 = 2$.
%    The hyperparameters in variational distributions of global variables are 
    The networks are trained with minibatch size $|B|$ of 128, momentum 0.9 and learning rate $1\times10^{-3}$.
    
    \paragraph{CIFAR-10} We set latent dim $d=64$ and $K=10$ here.
    The $p(\bo|\bx;\bgamma)$, $q(\bx|\bz, \bo)$ are both parameterized by 5 deep stack of residual blocks~\cite{he2016deep}, making the scale of images changing from $32\times32$ to $8\times 8$ and reversely for $q(\bx|\bz, \bo)$. The network of $q(\bz|\bo)$ is a simple 9-layer convolutional neural networks as adopted by recent deep semi-supervised methods~\cite{Luo_2018_CVPR}. The networks are trained for 200 epochs with Adam optimizer with learning rate $3\times 10^{-3}$ and minibatch size $|B| =100$ and $|S| = 200$.
    
    \paragraph{Crowdsourcing scheme} For both MNIST and CIFAR-10, we random sample a subset of 4000 images from the training dataset and ask workers on Amazon Mechanical Turks (AMT) to annotate them. In each task, 36 images are sampled from the subset and shown to the worker. Each worker is asked to divide 36 images into any number of clusters. From these feedbacks, we can derive pairwise annotations for following experiments.
    
    \section{Addtional experimental results}
\begin{figure}
        \centering
        \includegraphics[width=0.5\linewidth]{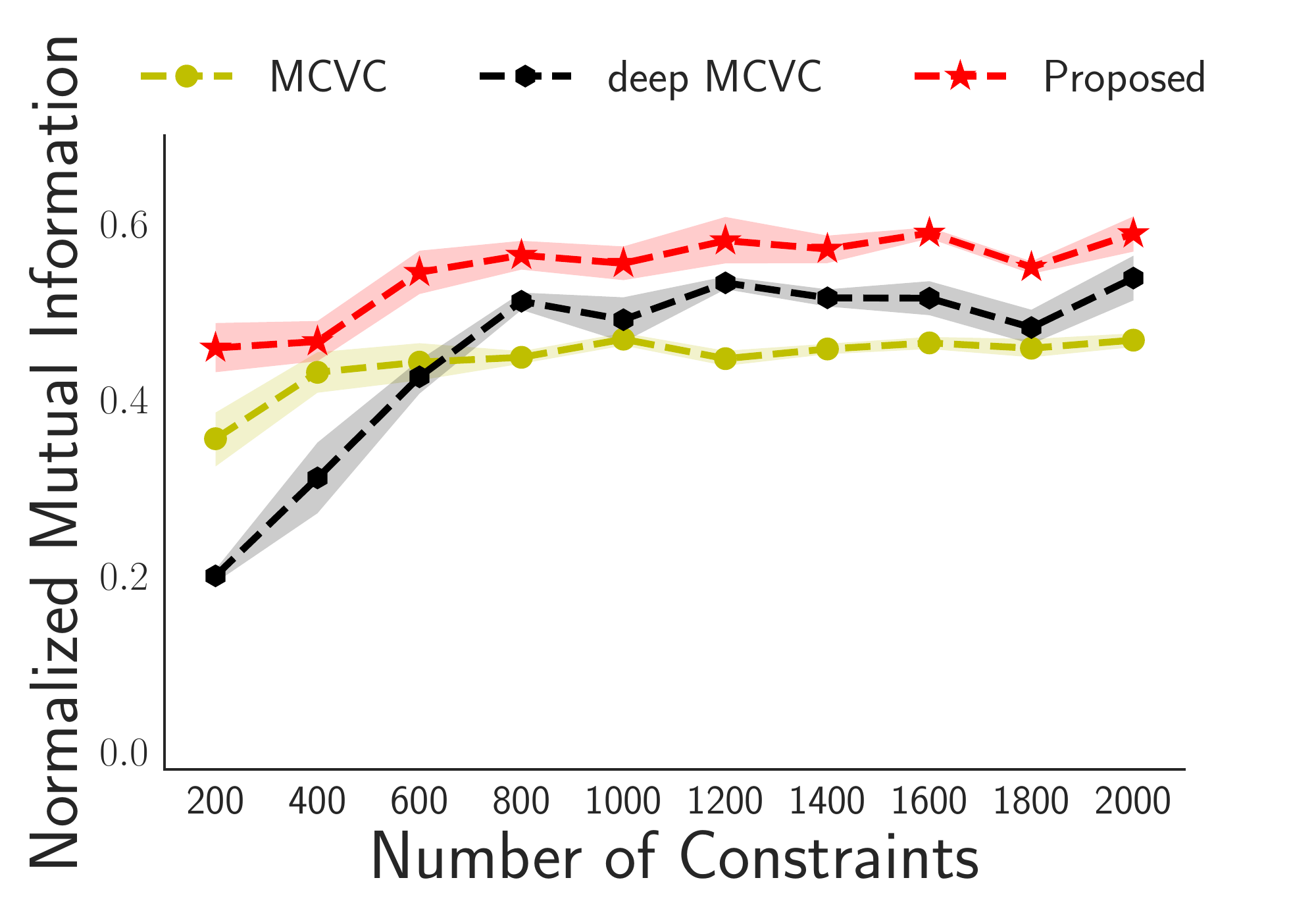}
        \caption{Comparison to deep MCVC on Face dataset with only a subset of 100 data points annotated. The experiment setting is the same as that in Fig.~\ref{fig:face100}.}
        \label{fig:pose100exp32unequalplotcomparewithdeepmcvcsmall}
    \end{figure}

    \paragraph{Comparison with deep MCVC~\cite{chang2017multiple}} In \Cref{sec:uci}, our model performs better than other competing methods because of the generative ``instance model" and thus the usage of structure information in unlabeled data. Discriminative methods such as MCVC~\cite{chang2017multiple} cannot take advantage of the observations of those unannotated data, even with a more complex deep neural network. %``a state of the art DL network".
    For fair comparison, we modify MCVC by using a more complex DNN (abbr. as deep MCVC). We test deep MCVC on the Face dataset, where deep MCVC uses the same network architecture (MLP with two hidden layers of 40 ReLU units) as our $q(\bz|\bo)$ in our method. As shown in Fig.~\ref{fig:pose100exp32unequalplotcomparewithdeepmcvcsmall}, the performance of deep MCVC is better than MCVC when more constraints are observed but worse than the proposed SCDC. The reason is that deep MCVC does not exploit the information of unannotated images during the training. 
    
    Furthermore, we also evaluate deep MCVC on CIFAR-10, whose best NMI score among 5 runs is 0.44. In contrast, the proposed SCDC achieves NMI score 0.55. The gap of the improvement is due to the help of unannotated data.
    %Also, we don't agree that "an estimation of z" can be easily obtained from the pairwise annotations. So other semi-supervised learning approaches can hardly be applied.
    
%    \paragraph{A third model between SCDC and BayesSCDC}

\end{document}